%% file: main.tex
\documentclass[11pt]{article}

\IfFileExists{acl.sty}{\usepackage[final]{acl}}{\usepackage[final]{latex/acl}}

\usepackage{times}
\usepackage{latexsym}
\usepackage[T1]{fontenc}
\usepackage[utf8]{inputenc}
\usepackage{microtype}
\usepackage{inconsolata}

\usepackage{amsmath}
\usepackage{amssymb}
\usepackage{graphicx}
\usepackage{booktabs}
\usepackage{array}
\usepackage{float}
\usepackage{adjustbox}
\usepackage{makecell}
\usepackage{placeins}
\usepackage{tikz}
\usetikzlibrary{arrows.meta,positioning}

\graphicspath{{figures/}}

\title{Evidence Absence Is Not Evidence Insufficiency: Diagnosing NEI Construction Artifacts in Fact Verification}

\author{
  \textbf{Jingxi Qiu\textsuperscript{1,2}},
  \textbf{Zeyu Han\textsuperscript{2}},
  \textbf{Cheng Huang\textsuperscript{1,\dag}}
  \\
  \textsuperscript{1}ZenWeave AI,
  \textsuperscript{2}Georgetown University,
  \textsuperscript{\dag}Corresponding Author
  \\
  \small{
    \href{mailto:jingxi@zenweaveai.com,chenghuang@zenweaveai.com}{\{jingxi,chenghuang\}@zenweaveai.com}
  }
}

\begin{document}
\maketitle

\input{sections/00_abstract}

\input{sections/01_introduction}
\input{sections/02_related_work}
\input{sections/03_neicap}
\input{sections/04_data}
\input{sections/05_results}
\input{sections/06_analysis}
\input{sections/07_conclusion}
\input{sections/08_limitations}

\bibliography{custom}

\newpage

\FloatBarrier
\appendix

\FloatBarrier
\input{appendices/A_NEI_CAP_Construction_Taxonomy}

\FloatBarrier
\input{appendices/B_Dataset_Construction_and_Manifests}

\FloatBarrier
\input{appendices/C_Artifact_Audit_Statistics}

\FloatBarrier
\input{appendices/D_Human_Validation_Protocol_and_Adjudication}

\FloatBarrier
\input{appendices/E_Experimental_Setup_and_Model_Details}

\FloatBarrier
\input{appendices/F_Full_SciFact_Construction_Matrices}

\FloatBarrier
\input{appendices/G_Human_Validated_Hard_NEI_Evaluation}

\FloatBarrier
\input{appendices/H_Multi_Model_Robustness}

\FloatBarrier
\input{appendices/I_Same_Claim_and_Same_Document_Diagnostics}

\FloatBarrier
\input{appendices/J_External_Controls_FEVER_and_HoVer}

\FloatBarrier
\input{appendices/K_Diagnostic_Case_Studies}

\FloatBarrier
\input{appendices/L_Statistical_Testing_and_Uncertainty_Estimation}

\FloatBarrier
\input{appendices/M_Reproducibility_Release_and_Claim_Boundary_CheckLIST}

\FloatBarrier
\input{appendices/N_Decoder_SFT_Diagnostics}

\end{document}

%% file: sections/00_abstract.tex
\begin{abstract}
Evidence absence is not evidence insufficiency, but fact verification benchmarks can make them observationally similar. The \textsc{Not Enough Information} (\textsc{NEI}) label is often operationalized through constructed evidence conditions, and that choice silently determines what a verifier learns. We introduce NEI-CAP, a construction-aware diagnostic protocol for insufficient-evidence evaluation. Each \textsc{NEI} example carries the construction family that produced it; NEI-CAP audits shortcut cues, validates hard cases through human adjudication, and tests whether competence transfers across constructions. We instantiate the protocol on SciFact, with FEVER and HoVer as bounded external controls. Across these settings, \textsc{NEI} competence does not transfer reliably: encoder verifiers and an instruction-tuned decoder trained on shortcut-prone constructions fail to recognize semantically related insufficient evidence, and mixed-construction training narrows but does not close the gap. Fixed-claim diagnostics further show that the evidence condition shifts confidence in the reference \textsc{Support}/\textsc{Refute} label, not only \textsc{NEI} recall, so an aggregate \textsc{NEI} score can hide which problem a model has actually solved. We therefore recommend reporting the construction family alongside the score, and distill the results into a checklist for benchmarks that carry an insufficient-evidence label.
\end{abstract}

%% file: sections/01_introduction.tex
\section{Introduction}
\label{sec:introduction}

A fact verification system labels a claim as supported, refuted, or \textsc{Not Enough Information} (\textsc{NEI}) when the available evidence is inconclusive~\citep{thorne2018fever,wadden2020scifact,jiang2020hover}. The \textsc{NEI} label is meant to be evidence-conditioned: for a claim $c$ and an evidence set $E$, it means that $E$ does not establish $c$ either way. Building those negative evidence sets, however, is itself a design step that no formal definition covers. An empty field, an off-topic passage, a high-overlap retrieval miss, and a non-rationale sentence drawn from a cited document can all be labeled \textsc{NEI}, and a verifier trained on one of these constructions can predict \textsc{NEI} for reasons that have little to do with whether the evidence is actually sufficient.

Prior work on fact-verification artifacts has focused on the claim side. \citet{schuster2019towards} show that FEVER can be partially solved by claim-only classifiers, and adversarial or contrastive verification resources further show that standard evidence-aware accuracy can hide brittle decision rules~\citep{thorne2019fever2,schuster2021vitaminc}. A separate line studies evidence sufficiency by removing parts of otherwise-valid evidence~\citep{atanasova2022fact}. We study a complementary failure mode---how the negative evidence condition is built in the first place---and argue that this construction silently determines what a verifier learns and what an aggregate \textsc{NEI}-F1 can hide.

\begin{figure*}[t]
    \centering
    \includegraphics[width=0.98\textwidth]{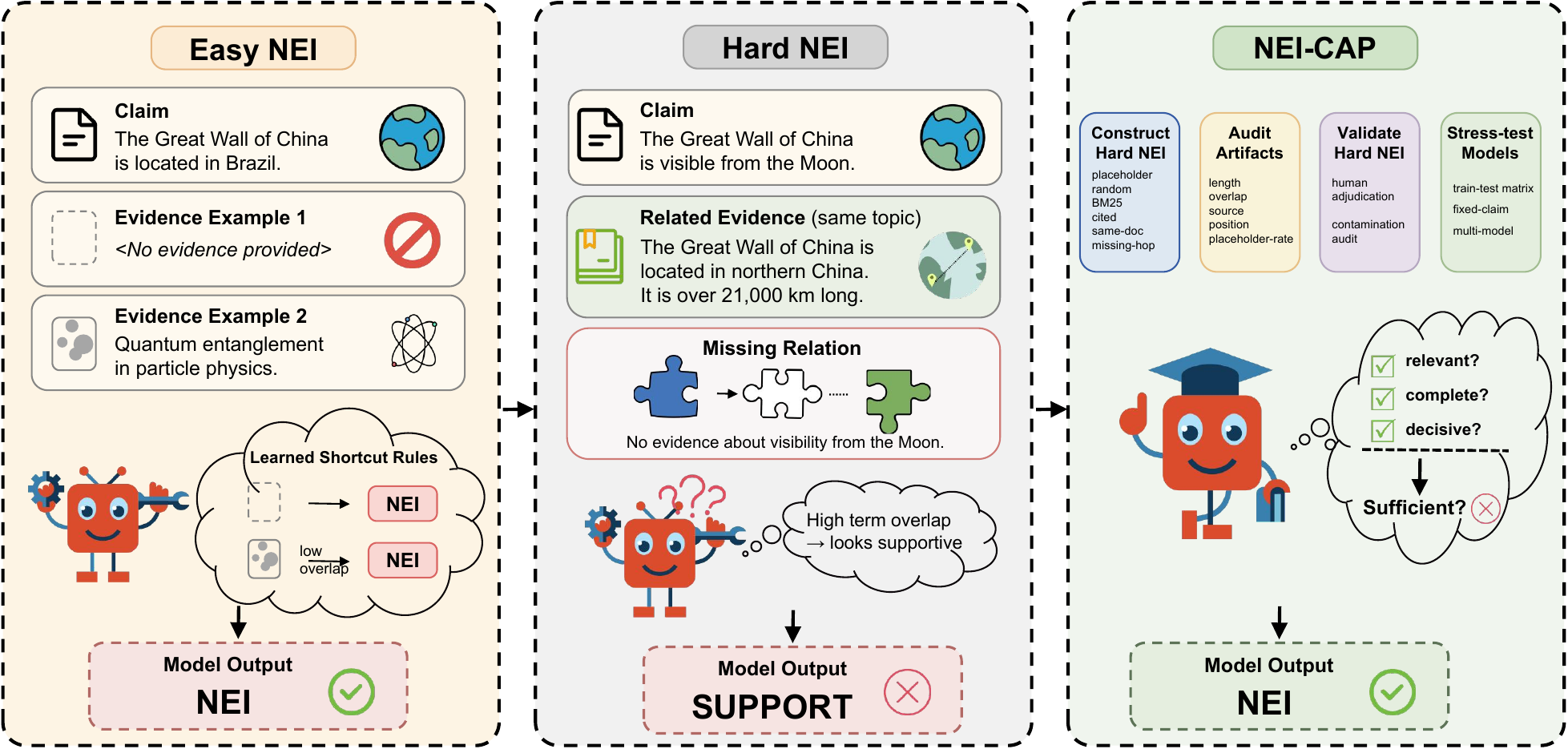}
    \caption{Conceptual illustration of \textsc{NEI} construction artifacts. Easy \textsc{NEI} constructions, such as placeholders or unrelated passages, can teach absence, format, or topic-mismatch shortcuts. Hard \textsc{NEI} keeps the evidence semantically related but incomplete; a verifier may therefore overpredict \textsc{Support} from overlap rather than recognize insufficiency. NEI-CAP records the construction family, audits shortcuts, validates hard examples, and stress-tests whether \textsc{NEI} competence transfers. Examples are schematic; experiments use the SciFact, FEVER, and HoVer constructions described in Sections~\ref{sec:neicap}--\ref{sec:data-setup}.}
    \label{fig:nei-construction-mechanism}
\end{figure*}

Figure~\ref{fig:nei-construction-mechanism} illustrates the mechanism. Easy \textsc{NEI} can be solved by recognizing absence, format, or topic mismatch; hard \textsc{NEI} keeps the evidence related to the claim but incomplete, which can induce false \textsc{Support} when the model mistakes overlap for sufficiency. NEI-CAP targets this gap by making construction family part of the evaluation record rather than treating \textsc{NEI} as a construction-free class.

NEI-CAP makes the construction explicit. Each \textsc{NEI} example carries the family of evidence condition that produced it, which lets us audit each family for shortcut features and stress-test whether a model trained on one family recognizes insufficiency in another. Across these variants, only the \textsc{NEI} evidence condition changes; the label space and claim distribution stay the same, so the comparisons measure construction sensitivity rather than domain transfer. We instantiate the protocol on SciFact-style scientific verification~\citep{wadden2020scifact}, with FEVER~\citep{thorne2018fever}, HoVer~\citep{jiang2020hover}, and the broader fact-verification literature as context~\citep{augenstein2019multifc,aly2021feverous}.

The headline finding is a transfer failure that aggregate \textsc{NEI}-F1 cannot detect. A DeBERTa verifier trained on placeholder \textsc{NEI} reaches perfect matched-placeholder \textsc{NEI}-F1 across five seeds, yet scores zero \textsc{NEI}-F1 on BM25 near-miss and cited non-rationale evaluation; the collapse replicates on RoBERTa and SciBERT. Probability mass shifts to \textsc{Support} and \textsc{Refute} rather than to \textsc{NEI}, so the failure reflects high-confidence misclassification rather than abstention. Training on random-irrelevant \textsc{NEI} fares only marginally better, showing that the problem extends beyond placeholder detection. Mixed-construction training narrows but does not close the gap. The failure is not specific to encoders. An instruction-tuned decoder handles hard \textsc{NEI} well zero-shot, but after fine-tuning on placeholder \textsc{NEI} alone it collapses in the same way, in every seed; fine-tuning on mixed constructions avoids most of the damage. A fixed-claim diagnostic further shows that swapping the evidence shifts confidence in the reference \textsc{Support} or \textsc{Refute} label, not only \textsc{NEI} recall. The construction choice therefore affects the verifier on the full three-way task, not only the \textsc{NEI} corner.

We make three contributions. \textit{First}, we recast \textsc{NEI} as a construction-sensitive evidence condition rather than a single negative label. \textit{Second}, we introduce NEI-CAP: a diagnostic protocol that treats the construction family as an explicit evaluation variable, audits its shortcut surface, and validates hard cases through human adjudication. \textit{Third}, our SciFact, FEVER, and HoVer experiments show that shortcut-prone training fails to transfer to semantically related insufficient evidence, and that multi-seed and mixed-construction protocols do not remove the need for construction-stratified reporting.

%% file: sections/02_related_work.tex
\section{Related Work}
\label{sec:related-work}

\subsection{Fact Verification Benchmarks}
Fact verification is commonly formulated as a three-way classification problem: given a claim and an evidence set, predict whether the evidence supports, refutes, or is insufficient to verify the claim. FEVER introduced a large-scale Wikipedia benchmark with \textsc{Supported}, \textsc{Refuted}, and \textsc{NotEnoughInfo} labels~\citep{thorne2018fever}; SciFact extended the formulation to expert-written scientific claims that require retrieving evidence-containing abstracts and rationales~\citep{wadden2020scifact}; and HoVer added many-hop evidence retrieval, where verification can depend on facts spread across multiple Wikipedia articles~\citep{jiang2020hover}. MultiFC broadens fact-checking to real-world multi-domain claims~\citep{augenstein2019multifc}; HealthFC focuses on evidence-backed medical claims~\citep{vladika2023healthfc}; FEVEROUS adds structured table evidence~\citep{aly2021feverous}; VitaminC creates contrastive claim--evidence pairs that require sensitivity to small factual changes~\citep{schuster2021vitaminc}; AVeriTeC collects real-world fact-checks whose verdict set includes a not-enough-evidence category~\citep{schlichtkrull2023averitec}; and SciFact-Open extends scientific claim verification to an open corpus of abstracts~\citep{wadden2022scifactopen}. Rationale-centered resources further ask whether systems identify the evidence used to support predictions~\citep{deyoung2020eraser}. Across these benchmarks the \textsc{NEI} label is treated as a fixed third class, but how its evidence side is built is left to each benchmark's discretion---and that choice is not made part of the evaluation protocol.

\subsection{Dataset Artifacts and Behavioral Evaluation}
NLP benchmarks routinely contain artifacts that let models score well without learning the intended capability, so aggregate scores can overstate progress~\citep{bowman2021will}. In natural language inference, hypothesis-only classifiers can recover the label from premise-free input~\citep{gururangan2018annotation,poliak2018hypothesis}, and controlled challenge sets such as HANS show that high benchmark accuracy can mask reliance on lexical or syntactic heuristics~\citep{mccoy2019right}. Similar shortcut effects appear beyond NLI, including argument reasoning artifacts~\citep{niven2019probing}; more broadly, shortcut learning is a known failure mode of modern neural systems~\citep{geirhos2020shortcut}. Multi-hop datasets show the same pattern: compositional questions can often be answered in a single hop~\citep{min2019compositional}, and models can exploit disconnected reasoning instead of combining facts across passages~\citep{trivedi2020multihop}. In fact verification, \citet{schuster2019towards} show analogous claim-side cues in FEVER and demonstrate that claim-only baselines remain competitive against evidence-aware models; FEVER2.0-style adversarial work further stresses robustness to perturbations~\citep{thorne2019fever2}. A broader line of work uses contrast sets, counterfactually augmented data, and behavioral testing to expose brittle shortcut reliance~\citep{kaushik2020learning,gardner2020evaluating,ribeiro2020checklist}. These studies mostly examine claim-side, hypothesis-side, or local decision-boundary artifacts. NEI-CAP carries the same diagnostic stance to the evidence side and asks what shortcuts the construction of \textsc{NEI} evidence itself can teach.

\subsection{Evidence Sufficiency and Missing Evidence}
A claim can be true or false in the world while the available evidence is still insufficient to settle it, so evidence sufficiency is a distinct question from veracity prediction. \citet{atanasova2022fact} make this question operational by removing parts of otherwise-valid evidence and asking whether fact-checking models notice the omission, and rationale evaluation benchmarks ask whether models identify supporting passages rather than only predicting labels~\citep{deyoung2020eraser}. Work on missing counter-evidence similarly argues that evidence availability and sufficiency are assumptions built into fact-checking datasets~\citep{glockner2022missing}. The same question appears in factuality evaluation for generation, where long-form claims are decomposed into atomic facts and checked against supporting sources~\citep{min2023factscore} and the granularity of that decomposition itself changes verification outcomes~\citep{lu2025optimizing}, and in retrieval-augmented generation or grounding evaluation, where generated claims must be supported by provided context~\citep{niu2024ragtruth,jacovi2025factsgrounding}. Sufficiency has also been studied on the retrieval side, where the question is whether the retrieved context alone could answer the query~\citep{joren2025sufficient}. It can further be treated as a property of the evidence set as a whole: missing hops and unresolved conflicts are not detectable when passages are scored independently~\citep{qiu2026surerag}. These studies motivate evidence-sensitive evaluation, but they usually treat the negative or unsupported condition as already given or derived from a valid one. NEI-CAP works in the opposite direction: it asks how the insufficient evidence set was built in the first place, and how that construction determines what a verifier can be said to have learned.

%% file: sections/03_neicap.tex
\section{NEI-CAP: Construction-Aware NEI Evaluation}
\label{sec:neicap}

Figure~\ref{fig:nei-construction-mechanism} motivates NEI-CAP as a way to separate shortcut recognition from evidence-insufficiency recognition. This section formalizes that idea with a construction variable, a compact taxonomy of evidence conditions, and the operational workflow in Protocol~1.

\subsection{Evidence-Conditioned NEI}
\label{subsec:evidence-conditioned-nei}

A verification instance is $(c,E,y)$, where $c$ is a claim, $E=\{e_1,\ldots,e_k\}$ is its evidence set, and $y\in\{\textsc{Support},\textsc{Refute},\textsc{NEI}\}$. The \textsc{NEI} label is a property of the pair $(c,E)$ rather than of the claim alone~\citep{thorne2018fever,wadden2020scifact}: whether the evidence is insufficient depends on what evidence is provided. We make this dependence explicit by extending each example with a construction variable,
\[
    x=(c,E,y,z,g),
\]
where $z$ records the family of \textsc{NEI} evidence condition that produced $E$ and $g$ is a grouping identifier that keeps variants of the same claim within a single split. The model never receives $z$ or $g$; they are diagnostic interventions, used only for auditing, splitting, and stratified reporting.

\subsection{NEI Construction Families}
\label{subsec:construction-families}

NEI-CAP separates shortcut-prone constructions from semantically related insufficient evidence. The former expose format, topic, position, or retrieval shortcuts; the latter test whether evidence that remains related to the claim is still recognized as insufficient. This follows the same motivation as contrast-set and behavioral evaluation: a benchmark should expose when a model succeeds through an unintended decision rule rather than the intended capability~\citep{gardner2020evaluating,ribeiro2020checklist,geirhos2020shortcut}. We keep five evidential roles distinct: shortcut anchors, candidate hard constructions, human-adjudicated hard constructions, paired diagnostics, and external controls. Each result is reported with the family that produced it. Table~\ref{tab:nei-taxonomy} lists the compact taxonomy used in the rest of the paper; full definitions, metadata fields, and shortcut-risk dimensions are in Appendix~\ref{app:taxonomy}.

\begin{table}[t]
\centering
\small
\begin{adjustbox}{max width=\columnwidth}
\begin{tabular}{@{}l>{\raggedright\arraybackslash}p{0.30\columnwidth}>{\raggedright\arraybackslash}p{0.26\columnwidth}@{}}
\toprule
\textbf{Family} & \textbf{Evidence condition} & \textbf{Role} \\
\midrule
Placeholder & Fixed/empty no-evidence marker & Format shortcut anchor \\
Random irrelevant & Unrelated evidence & Topic-mismatch anchor \\
Position-biased & Predictable non-rationales & Position/source audit \\
BM25 near-miss & High-overlap insufficient evidence & Hard \textsc{NEI} \\
Cited non-rationale & Cited but non-rationale evidence & Hard \textsc{NEI} \\
Same-document & Same-source non-rationale evidence & Source-controlled \textsc{NEI} \\
Fixed-claim & Same claim, changed evidence & Evidence-substitution diagnostic \\
Missing-hop & Multi-hop evidence, required fact removed & External multi-hop control \\
\bottomrule
\end{tabular}
\end{adjustbox}
\caption{Compact NEI-CAP construction taxonomy.}
\label{tab:nei-taxonomy}
\end{table}

\subsection{Diagnostic Protocol}
\label{subsec:diagnostic-protocol}

Protocol~1 lists the five stages that produce the construction-stratified evidence reported in Sections~\ref{sec:results}--\ref{sec:analysis}. The stages share a common output discipline: each one returns a typed artifact that the next stage can consume without re-deriving anything from the raw text.

\begin{center}
\small
\setlength{\fboxsep}{5pt}
\setlength{\fboxrule}{0.45pt}
\fbox{%
\begin{minipage}{0.93\columnwidth}
\raggedright
\textbf{Protocol 1: NEI-CAP diagnostic workflow}\\[-1pt]
\rule{\linewidth}{0.3pt}
\textbf{Input:} claim--evidence examples $(c,E,y)$ and construction rules.\\
\textbf{Output:} audit tables, adjudicated subsets, construction-stratified metrics, and claim boundaries.\\[2pt]
\textbf{1. Construct.} Assign construction family $z$ and group ID $g$.\\
\textbf{2. Audit.} Measure evidence-side shortcut features by label and construction.\\
\textbf{3. Validate.} Adjudicate candidate hard \textsc{NEI} used for central claims.\\
\textbf{4. Stress-test.} Evaluate single- and mixed-construction training under construction-stratified tests.\\
\textbf{5. Report.} Release construction-specific metrics, uncertainty, and claim boundaries.
\end{minipage}%
}
\end{center}

%% file: sections/04_data.tex
\section{Data, Validation, and Experimental Setup}
\label{sec:data-setup}

SciFact is our primary setting~\citep{wadden2020scifact}: it requires evidence-sensitive verification over scientific abstracts, where reference evidence and semantically related but insufficient evidence routinely coexist in the same document. FEVER and HoVer serve as bounded external controls~\citep{thorne2018fever,jiang2020hover}.

\subsection{SciFact Construction Suite}
\label{subsec:scifact-suite}

The main SciFact suite instantiates the construction families in Table~\ref{tab:nei-taxonomy}: placeholder, random irrelevant, position-biased, BM25 near-miss, and cited non-rationale. BM25 near-miss examples are obtained with BM25 retrieval \citep{robertson2009bm25} and then filtered to retain high-overlap but insufficient evidence. The \textsc{Support} and \textsc{Refute} portions of the task are held comparable across variants---every test variant contains 76 \textsc{Support}, 40 \textsc{Refute}, and 61 \textsc{NEI} examples---while only the \textsc{NEI} evidence condition changes. This lets us train a verifier under one \textsc{NEI} construction and evaluate it under another. Splits are group-disjoint, keyed by the original claim or claim--document grouping, so construction variants of the same claim never straddle train and test partitions.

\subsection{Human-Adjudicated Hard NEI}
\label{subsec:human-validation}

Candidate hard \textsc{NEI} examples are useful only if they are actually insufficient: BM25 near-miss, cited non-rationale, and same-document non-rationale evidence can carry implicit support or refutation that the construction rule does not catch. NEI-CAP therefore separates \emph{candidate} hard \textsc{NEI} from \emph{human-adjudicated} hard \textsc{NEI}.

We use two validation resources (Table~\ref{tab:human-validation-summary}). The SciFact hard-\textsc{NEI} audit adjudicates candidate BM25/cited near-miss examples from the construction suite; the fixed-claim/same-document audit covers examples where the claim or source document is held fixed while the evidence condition changes. Two annotators produced consensus adjudications; five-way primary-label agreement is 94.4\% (Cohen's $\kappa=0.732$), and agreement on the valid-\textsc{NEI}-versus-contamination gate that defines the hard subsets is 94.8\% ($\kappa=0.743$). AI-assisted checks were used only as secondary support. The full protocol, label schema, and agreement analysis are in Appendix~\ref{app:human-validation}.

The primary human-hard model evaluation uses the held-out test split of the SciFact human-adjudicated hard-\textsc{NEI} audit: 54 validated hard-\textsc{NEI} examples after group-disjoint splitting. The larger audit pool estimates label validity, while the held-out subset supports model evaluation. Appendix~\ref{app:human-hard-eval} reports the full evaluation table and Appendix~\ref{app:data} documents the sampling trail.

\begin{table}[H]
\centering
\small
\begin{adjustbox}{max width=\columnwidth}
\begin{tabular}{lrrrr}
\toprule
\textbf{Validation resource} & \textbf{$n$} & \textbf{Valid NEI} & \textbf{Contam.} & \textbf{Hard} \\
\midrule
BM25/cited hard-\textsc{NEI} audit & 250 & 89.2\% & 10.8\% & 195 (87.4\%) \\
Fixed-claim/same-doc audit         & 122 & 96.7\% & 3.3\%  & 114 (96.6\%) \\
\bottomrule
\end{tabular}
\end{adjustbox}
\caption{Human validation summary. Contamination includes examples adjudicated as actually supported, actually refuted, ambiguous, or invalid. The Hard column reports the absolute number of human-adjudicated hard \textsc{NEI} examples that pass the audit, with the in-parentheses rate computed against valid \textsc{NEI}. Full adjudication details are in Appendix~\ref{app:human-validation}.}
\label{tab:human-validation-summary}
\end{table}

\subsection{Models and Metrics}
\label{subsec:models-metrics}

Our primary verifier is DeBERTa-v3-base~\citep{he2021debertav3}, with RoBERTa-base and SciBERT as secondary backbones~\citep{liu2019roberta,beltagy2019scibert}. We use these as diagnostic probes rather than proposed architectures: training variants differ only in the \textsc{NEI} construction family used during training. Mixed-construction regimes combine the single-construction variants into an easy mixture (placeholder, random irrelevant, position-biased), a hard mixture (BM25 near-miss, cited non-rationale), and a balanced mixture over all five families. Multi-seed experiments use seeds 13, 17, 23, 29, and 37. Qwen2.5-7B-Instruct~\citep{yang2024qwen25} serves as a decoder probe, evaluated zero-shot and under controlled five-seed LoRA fine-tuning; its design is described in Section~\ref{subsec:decoder-results} and Appendix~\ref{app:decoder-sft}.

For three-way verification, we report Macro-F1 and class-specific F1, especially \textsc{NEI}-F1. For human-validated hard-\textsc{NEI} subsets, every example is adjudicated as \textsc{NEI}, so Macro-F1 is not informative; we instead report \textsc{NEI} recall, false \textsc{Support} rate, false \textsc{Refute} rate, and mean predicted class probabilities. Hyperparameters, checkpoint selection, seed aggregation, and metric details are in Appendices~\ref{app:setup} and~\ref{app:statistics}.

\subsection{External Controls}
\label{subsec:external-controls}

FEVER and HoVer are Wikipedia verification datasets; all core experiments run on SciFact. FEVER has a native \textsc{NotEnoughInfo} label~\citep{thorne2018fever} and serves as a non-toy subset control. HoVer requires multi-hop evidence, so sufficiency can depend on more than one supporting fact~\citep{jiang2020hover}; it serves as a candidate missing-hop control. Full details are in Appendix~\ref{app:external-controls}.

%% file: sections/05_results.tex
\section{Results}
\label{sec:results}

\subsection{Matched NEI Performance Does Not Imply Transfer}
\label{subsec:scifact-construction-sensitivity}

\begin{figure}[H]
    \centering
    \includegraphics[width=\columnwidth]{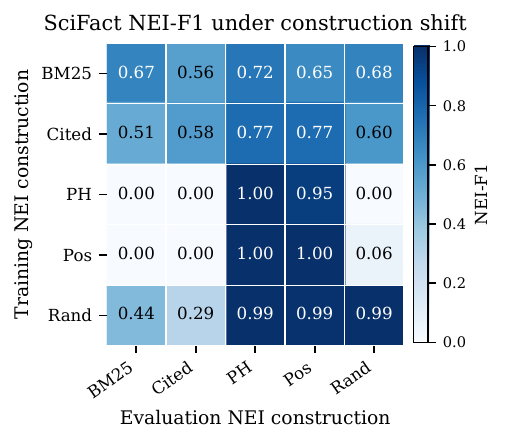}
    \caption{SciFact \textsc{NEI}-F1 under train/test construction shifts. Cell values are seed-aggregated means from the primary construction matrix; Appendix~\ref{app:scifact-matrices} reports full numeric matrices and Macro-F1. Matched easy \textsc{NEI} can yield high scores without transferring to semantically related insufficient evidence.}
    \label{fig:scifact-matrix}
\end{figure}

Figure~\ref{fig:scifact-matrix} shows the SciFact train/test construction matrix. A placeholder-trained verifier obtains perfect matched-placeholder \textsc{NEI}-F1, but falls to 0.000 on BM25 near-miss and cited non-rationale \textsc{NEI}. Position-biased training shows the same hard-construction collapse, while random-irrelevant training is less extreme but still shortcut-prone: it is nearly solved under matched evaluation yet transfers poorly to BM25 and cited hard constructions. Full numeric matrices are in Appendix~\ref{app:scifact-matrices}.

The point is not that placeholder evidence is realistic; it is an intentionally shortcut-prone anchor. The broader finding is that different easy constructions teach different shortcuts---format absence, position/source bias, or topic mismatch---whereas the target of NEI-CAP is hard \textsc{NEI}: semantically related evidence that remains insufficient.

\subsection{Human-Adjudicated Hard NEI Exposes Easy-Training Failure}
\label{subsec:human-hard-results}

Human validation does not rescue easy \textsc{NEI} training. Table~\ref{tab:human-hard-nei-results} shows that placeholder and position-biased training yield 0.000 recall on human-adjudicated hard \textsc{NEI}. Random-irrelevant training also performs poorly (0.216 recall), confirming that topic-unrelated \textsc{NEI} does not teach hard-insufficiency recognition. Errors are dominated by false \textsc{Support}, and BM25 near-miss and cited non-rationale training partially recover recall.

\begin{table}[H]
\centering
\small
\begin{adjustbox}{max width=\columnwidth}
\begin{tabular}{lrrr}
\toprule
\textbf{Train NEI} & \textbf{Recall} & \textbf{False SUP} & \textbf{False REF} \\
\midrule
BM25 near-miss & 0.691 & 0.142 & 0.167 \\
Cited non-rationale & 0.679 & 0.074 & 0.247 \\
Random irrelevant & 0.216 & 0.574 & 0.210 \\
Placeholder & 0.000 & 0.870 & 0.130 \\
Position-biased & 0.000 & 0.907 & 0.093 \\
\bottomrule
\end{tabular}
\end{adjustbox}
\caption{Evaluation on the held-out SciFact human-adjudicated hard-\textsc{NEI} test split ($n=54$). Since all examples are validated as \textsc{NEI}, we report \textsc{NEI} recall and error rates rather than Macro-F1; Appendix~\ref{app:data} gives the sampling trail from the larger audit pool.}
\label{tab:human-hard-nei-results}
\end{table}

Placeholder training yields 0/54 recall, with a Clopper--Pearson 95\% upper bound of 0.066~\citep{clopper1934use}; BM25 near-miss training gives a bootstrap interval of $[0.593, 0.790]$. Under placeholder training, mean $P(\textsc{NEI})$ on the validated hard examples is 0.013, with the remaining probability mass on \textsc{Support} and \textsc{Refute}, so the failure is high-confidence misclassification rather than abstention (Appendix~\ref{app:human-hard-eval}).

\subsection{Mixed Training Helps but Remains Construction-Stratified}
\label{subsec:mixed-training-results}

Mixed training is a stronger test because real benchmarks rarely contain a single \textsc{NEI} construction. Table~\ref{tab:mixed-training-summary} summarizes DeBERTa under an easy mixture, a hard BM25/cited mixture, and a balanced all-family mixture. Easy-mixture training remains weak on hard constructions; hard and balanced mixtures improve recall but still produce construction-stratified profiles. Mixed training changes the failure profile---it does not make \textsc{NEI} construction-free.

\begin{table}[H]
\centering
\small
\begin{adjustbox}{max width=\columnwidth}
\begin{tabular}{lrrrp{0.23\linewidth}}
\toprule
\textbf{Train regime} & \textbf{BM25} & \textbf{Cited} & \textbf{Hard recall} & \textbf{Takeaway} \\
\midrule
Easy mixture & 0.379 & 0.273 & 0.178 & Shortcut-heavy mixture remains weak \\
Hard mixture & 0.662 & 0.652 & 0.770 & Hard training improves recall \\
Balanced mixture & 0.802 & 0.803 & 0.915 & Best, but still stratified \\
\bottomrule
\end{tabular}
\end{adjustbox}
\caption{DeBERTa mixed-construction training summary. BM25 and Cited are construction-stratified \textsc{NEI}-F1; Hard recall is one-class recall on the SciFact human-adjudicated hard-\textsc{NEI} subset. Easy mixture includes random-irrelevant and position-biased \textsc{NEI}, so it improves over placeholder-only training but remains weak on hard insufficient evidence.}
\label{tab:mixed-training-summary}
\end{table}

The balanced mixture improves the hard-side metrics, but the full stratified matrix in Appendix~\ref{app:multimodel} still shows different performance profiles across placeholder, random-irrelevant, BM25, and cited conditions; mixed training is therefore a stronger stress test, not a replacement for construction-aware reporting.

\subsection{Placeholder-to-Hard Collapse Is Stable Across Seeds and Backbones}
\label{subsec:multimodel-results}

The placeholder-to-hard collapse is not specific to DeBERTa or to one seed. Across five seeds for DeBERTa, RoBERTa, and SciBERT, each backbone reaches perfect matched-placeholder \textsc{NEI}-F1 and zero \textsc{NEI}-F1 on BM25 near-miss and cited non-rationale evaluation when trained on placeholder \textsc{NEI} (Table~\ref{tab:multimodel-drop}). That the same exact 1.000$\rightarrow$0.000 pattern reproduces across three encoder backbones and fifteen training runs rules out interpreting the collapse as a training-dynamic artifact of any one backbone.

\begin{table}[H]
\centering
\small
\begin{adjustbox}{max width=\columnwidth}
\begin{tabular}{lrrrrr}
\toprule
\textbf{Model} & \textbf{Seeds} & \textbf{PH$\rightarrow$PH} & \textbf{PH$\rightarrow$BM25} & \textbf{PH$\rightarrow$Cited} & \textbf{Drop} \\
\midrule
DeBERTa & 5 & 1.000 & 0.000 & 0.000 & 1.000 \\
RoBERTa & 5 & 1.000 & 0.000 & 0.000 & 1.000 \\
SciBERT & 5 & 1.000 & 0.000 & 0.000 & 1.000 \\
\bottomrule
\end{tabular}
\end{adjustbox}
\caption{Five-seed placeholder-to-hard robustness. PH denotes placeholder \textsc{NEI}. Scores of 0.000 are exact: in all five seeds, placeholder-trained models never predicted \textsc{NEI} on the BM25 or cited hard test sets, yielding zero \textsc{NEI}-F1.}
\label{tab:multimodel-drop}
\end{table}

\subsection{Source-Controlled Diagnostics Are Useful but Bounded}
\label{subsec:same-claim-results}

The fixed-claim diagnostic is the closest available paired control: across 114 paired cases, the same claim is paired first with reference evidence and then with human-adjudicated insufficient evidence. Holding the claim fixed removes claim-side variation, so behavioral changes arise from evidence substitution; because the paired passages are not feature-matched, this diagnostic does not isolate semantic sufficiency alone. The probability that the verifier assigns to the reference \textsc{Support}/\textsc{Refute} label drops on the insufficient side; construction choice therefore affects how confidently the model commits to the non-\textsc{NEI} labels, not only whether it predicts \textsc{NEI}. Same-document hard \textsc{NEI} provides an additional source-controlled view that retains human adjudication; a shallow-feature audit shows it remains constructionally distinct from BM25/cited hard \textsc{NEI}, so we report it as its own diagnostic family rather than as a universal hard-\textsc{NEI} score. Full fixed-claim and same-document diagnostics are in Appendix~\ref{app:same-claim}, with row-level examples in Appendix~\ref{app:case-studies}.

\subsection{Zero-Shot Robustness Does Not Survive Placeholder Fine-Tuning}
\label{subsec:decoder-results}

We evaluate Qwen2.5-7B-Instruct~\citep{yang2024qwen25} zero-shot with deterministic conditional label scoring, then fine-tune it with LoRA~\citep{hu2022lora} across five seeds, changing only the training \textsc{NEI} realization between regimes and holding everything else fixed. Appendix~\ref{app:decoder-sft} reports the design, data controls, and per-regime details.

\begin{table}[H]
\centering
\small
\begin{adjustbox}{max width=\columnwidth}
\begin{tabular}{lrrrr}
\toprule
\textbf{Qwen regime} & \textbf{PH} & \textbf{Hard} & \textbf{Hum.\ recall} & \textbf{REF-F1} \\
\midrule
Zero-shot base & 0.847 & 0.793 & 0.963 & 0.532 \\
Placeholder-only SFT & 1.000 {\scriptsize$\pm$0.000} & 0.054 {\scriptsize$\pm$0.054} & 0.011 & 0.645 \\
Hard-mixture SFT & 0.920 {\scriptsize$\pm$0.013} & 0.834 {\scriptsize$\pm$0.008} & 0.900 & 0.784 \\
Balanced-all SFT & 0.955 {\scriptsize$\pm$0.012} & 0.817 {\scriptsize$\pm$0.018} & 0.856 & 0.795 \\
\bottomrule
\end{tabular}
\end{adjustbox}
\caption{Qwen2.5-7B decoder diagnostic. PH is placeholder-test \textsc{NEI}-F1; for the placeholder-only regime this equals the matched-construction score. Hard is mean \textsc{NEI}-F1 over BM25 near-miss and cited non-rationale evaluation; Hum.\ recall is \textsc{NEI} recall on the held-out human-adjudicated hard-\textsc{NEI} split ($n=54$). SFT rows are five-seed means with sample standard deviations.}
\label{tab:qwen-decoder-summary}
\end{table}

Zero-shot Qwen does not reproduce the encoder collapse (Table~\ref{tab:qwen-decoder-summary}), but its high human-hard recall comes partly from overpredicting \textsc{NEI}: it predicts \textsc{Support}/\textsc{Refute}/\textsc{NEI} at 45.4/9.3/45.3\%, against a gold distribution of 42.9/22.6/34.5\%, with \textsc{Refute}-F1 of 0.532. Placeholder-only SFT then induces the encoder-style failure in every seed: matched-placeholder \textsc{NEI}-F1 stays at 1.000 while hard \textsc{NEI}-F1 falls to 0.054 and human-hard recall to 0.011. Hard-mixture and balanced SFT recover most of the loss, and they do so by discriminating the three classes better, not by predicting \textsc{NEI} more often: balanced SFT moves the predicted distribution to 43.0/22.7/34.3\%, raises \textsc{Refute}-F1 to 0.795, and keeps hard \textsc{NEI}-F1 at 0.817.

A verifier that is robust when prompted can therefore lose that robustness during fine-tuning. Llama-3.1-8B-Instruct~\citep{grattafiori2024llama} failed our predeclared label-order stability gate and was not fine-tuned.

\subsection{External Controls Mirror the SciFact Pattern}
\label{subsec:external-control-results}

On FEVER and HoVer, a placeholder-trained baseline reaches \textsc{NEI}-F1 of 1.000 on placeholder evaluation and 0.000 on BM25 (FEVER) or missing-hop (HoVer) evaluation, mirroring the SciFact pattern at a different model scale and on a different domain. Full external-control results are in Appendix~\ref{app:external-controls}.

%% file: sections/06_analysis.tex
\section{Analysis and Discussion}
\label{sec:analysis}

\subsection{NEI Is a Family of Evidence Conditions}
\label{subsec:hard-nei-heterogeneity}

Hard \textsc{NEI} should not be treated as one uniform category. BM25 near-miss, cited non-rationale, same-document non-rationale, fixed-claim hard \textsc{NEI}, and missing-hop evidence remove different shortcuts and stress different behaviors. Construction-level audits show that these families differ in evidence length, claim--evidence overlap, coverage, and retrieval metadata (Appendix~\ref{app:audit}). These audits separate evidence-insufficiency evaluation from construction recognition: the families differ in surface features, so results are reported per family rather than folded into aggregate \textsc{NEI}-F1. Random irrelevant evidence is especially important: it can teach topic mismatch rather than insufficiency, demonstrating that the problem is broader than placeholder detection.

\subsection{Mixed Training Does Not Remove the Reporting Requirement}
\label{subsec:mixed-training-discussion}

Mixed-construction training does not remove construction sensitivity (Section~\ref{subsec:mixed-training-results}), so construction-aware reporting remains necessary even for models trained on diverse constructions. NEI-CAP's output is a reporting discipline rather than a better aggregate score: every result carries the construction family that produced it, the audit metrics that describe its shortcut surface, and the validation status of any hard examples it relies on.

\subsection{What Fixed-Claim Diagnostics Can and Cannot Show}
\label{subsec:same-claim-discussion}

Fixed-claim evidence substitution separates two abilities: recognizing semantically related evidence as insufficient, and assigning the reference label when decisive evidence is present. The probability-drop metric shows that replacing reference evidence with insufficient evidence lowers confidence in the reference label, so construction affects \textsc{Support}/\textsc{Refute} verification confidence as well as \textsc{NEI} recall.

Same-document hard \textsc{NEI} is human-adjudicated, yet shallow features nearly separate it from BM25/cited hard \textsc{NEI} and other construction families. Human validity and constructional distinctness are different properties: even human-valid hard \textsc{NEI} can carry construction-specific signatures.

\subsection{A Reporting Checklist for Practitioners}
\label{subsec:practitioner-checklist}

We distill the diagnostic results into a reporting checklist for any benchmark or system evaluation that includes an insufficient-evidence label. Where dataset documentation records provenance for a resource as a whole~\citep{gebru2021datasheets}, this checklist records it per \textsc{NEI} construction. A report should record: (1)~\emph{construction provenance}---the family of evidence condition that produced each \textsc{NEI} example; (2)~a \emph{shortcut audit} of evidence-side features by label and construction family; (3)~\emph{validation status and uncertainty}---which hard examples are human-adjudicated, with intervals when validated subsets are small; (4)~\emph{aggregate and stratified metrics}---per-family and worst-family \textsc{NEI} scores alongside any aggregate; (5)~\emph{false \textsc{Support} and false \textsc{Refute} rates}, since confidently accepting a claim on insufficient evidence is the central downstream risk; and (6)~the \emph{model and training regime}, because construction sensitivity can be training-induced even in models that look robust zero-shot (Section~\ref{subsec:decoder-results}). Protocol~1 operationalizes items~(1)--(4); Appendix~\ref{app:reproducibility} packages the checklist together with claim boundaries.

Family selection follows the same logic as the diagnosis: the \textsc{NEI} constructions in a benchmark should mirror the negative evidence conditions its retrieval setting actually produces. In SciFact-style scientific verification, related-but-insufficient abstracts motivate the BM25 near-miss and cited non-rationale families; in multi-hop settings such as HoVer, incomplete evidence chains motivate missing-hop constructions. Whatever the domain, the results support a minimum: one shortcut-prone anchor, kept as an artifact control rather than as competence evidence, and one semantically related hard family validated by human adjudication, since these examples carry the competence claims.

Because \textsc{NEI} is evidence-conditioned, high-stakes deployment calls for construction-stratified regression testing, uncertainty communication, and human escalation paths, not a single aggregate acceptance score.

%% file: sections/07_conclusion.tex
\section{Conclusion}
\label{sec:conclusion}

\textsc{NEI} is often treated as a single negative label, but the way it is built decides what a verifier can actually learn. In SciFact-style verification, a placeholder-trained verifier earns a perfect matched-placeholder \textsc{NEI}-F1 yet scores zero on evidence that is topically related but insufficient---a pattern that reproduces across three encoder backbones, five seeds, and two external datasets.

NEI-CAP makes the construction explicit. By attaching every \textsc{NEI} example to the family of evidence condition that produced it and validating the hard cases through human adjudication, the protocol turns insufficient-evidence evaluation from a single number into a construction-stratified report.

The construction of insufficient evidence belongs alongside the labels and metrics that define any fact-verification benchmark. An aggregate \textsc{NEI}-F1 that hides how its \textsc{NEI} examples were built cannot tell a reader whether the model has learned to recognize insufficiency or simply to recognize the artifact.

%% file: sections/08_limitations.tex
\section*{Limitations}
\label{sec:limitations}

NEI-CAP is a diagnostic protocol. Strong results under it do not guarantee general-purpose sufficiency reasoning, and other domains may need additional construction families and annotation guidelines. The suite also favors internal validity: each construction family isolates one \textsc{NEI} evidence condition, whereas production retrieval mixes missing, irrelevant, partial, stale, and conflicting evidence.

Our strongest evidence comes from SciFact-style scientific verification. FEVER and HoVer show that the placeholder-to-hard pattern is not unique to SciFact, but they are narrower designs: the construction matrix, human adjudication, and training experiments were not repeated on them. Multi-domain expert adjudication of hard \textsc{NEI} remains future work.

The held-out human-adjudicated split is small ($n=54$). Adjudication is expensive, and the audit pipeline shrinks the pool at every stage: 250 candidates, 223 adjudicated valid \textsc{NEI}, 195 hard, and 54 after group-disjoint splitting. At this size the split can separate large effects, but it cannot rank insufficiency subtypes: 54 examples spread over broad-topic, near-miss, and partial subtypes leave too few per subtype. It also cannot estimate subtype prevalence, because the audit pool was selected for hardness, not sampled to be representative. Cross-domain conclusions would require comparable audits outside SciFact.

Human validation reduces label-validity risk but does not eliminate it. Near-miss and non-rationale evidence can carry implicit support, refutation, or ambiguity that two annotators may both miss. Agreement was weakest on fine-grained insufficiency subtypes (54.9\% conditional agreement, $\kappa=0.363$): annotators reliably agreed on whether an example was hard, but often differed on which kind of insufficiency it exhibited, which is consistent with the ambiguity-driven disagreement documented for claim--evidence pairs by \citet{glockner2024ambifc}. Subtype-level analyses should therefore be read as descriptive (Appendix~\ref{app:human-validation}).

Fixed-claim and same-document diagnostics are decomposition tests. The paired evidence passages are not feature-matched: the reference and insufficient passages can differ in length, style, or specificity, so paired behavioral changes cannot be attributed to sufficiency alone. Same-document hard \textsc{NEI} also remains constructionally distinct under shallow-feature audits, and the matched length/coverage audit was too small to settle how much these surface differences matter.

Our model probes are encoder cross-encoders plus one decoder. The decoder evidence covers Qwen2.5-7B on SciFact under one adaptation method (LoRA); Llama-3.1-8B failed a predeclared label-order stability gate and was not fine-tuned, so we cannot say how far the training-induced failure extends across decoder families, scales, or adaptation methods. Extending NEI-CAP to generative factuality and grounding evaluation~\citep{min2023factscore,niu2024ragtruth,jacovi2025factsgrounding} is future work.

%% file: appendices/A_NEI_CAP_Construction_Taxonomy.tex
\section{NEI-CAP Construction Taxonomy}
\label{app:taxonomy}

This appendix expands the compact taxonomy in Table~\ref{tab:nei-taxonomy}. NEI-CAP treats \textsc{NEI} not as a construction-free negative label, but as an evidence-conditioned label whose interpretation depends on how the evidence set is paired with the claim.

\subsection{Taxonomy Axes}

We characterize each \textsc{NEI} construction family along six axes: evidence availability, topical relatedness, source control, lexical or entity overlap, evidential completeness, and validation status. These axes separate absence-based shortcuts from semantically related insufficient evidence.

\paragraph{Evidence availability.}
Some \textsc{NEI} examples contain no substantive evidence or use a fixed placeholder. Others contain non-empty evidence that is related to the claim but still insufficient.

\paragraph{Topical relatedness.}
Random irrelevant evidence may be insufficient because it is off topic. BM25 near-miss, cited non-rationale, same-document non-rationale, and missing-hop examples are instead related to the claim but incomplete.

\paragraph{Source control.}
Same-document non-rationales and fixed-claim diagnostics reduce source and topic shortcuts by holding the source document or claim fixed.

\paragraph{Lexical and entity overlap.}
Near-miss examples preserve lexical or entity overlap while removing decisive support, testing whether models mistake overlap for sufficiency.

\paragraph{Evidential completeness.}
Partial and missing-hop examples contain some relevant information while omitting a necessary relation, condition, or reasoning step.

\paragraph{Validation status.}
NEI-CAP distinguishes constructed examples, candidate hard \textsc{NEI}, and human-adjudicated hard \textsc{NEI}. The strongest hard-\textsc{NEI} claims are restricted to human-adjudicated subsets.

\subsection{Construction Families}

\begin{table}[H]
\centering
\scriptsize
\begin{adjustbox}{max width=\columnwidth}
\begin{tabular}{p{0.16\linewidth}p{0.26\linewidth}p{0.20\linewidth}p{0.16\linewidth}p{0.14\linewidth}}
\toprule
\textbf{Family} & \textbf{Construction rule} & \textbf{Shortcut risk} & \textbf{Validation status} & \textbf{Paper role} \\
\midrule
Placeholder &
Replace evidence with a fixed or empty no-evidence marker. &
Format, length, absence. &
Constructed only. &
Shortcut baseline. \\
Random irrelevant &
Pair claim with evidence from unrelated claims or documents. &
Topic mismatch, low overlap. &
Constructed only. &
Topic-mismatch baseline. \\
Position-biased &
Select non-rationale evidence from predictable locations. &
Sentence position, source distribution. &
Not primary human-validated. &
Artifact stress test. \\
BM25 near-miss &
Retrieve high-overlap evidence that remains insufficient. &
Lexical overlap mistaken for support. &
Human-adjudicated in SciFact audit. &
Hard \textsc{NEI}. \\
Cited non-rationale &
Use evidence from claim-associated cited documents that is not the rationale. &
Source relevance mistaken for sufficiency. &
Human-adjudicated in SciFact audit. &
Hard \textsc{NEI}. \\
Same-document non-rationale &
Use non-rationale evidence from the same source document. &
Residual topic/source cues. &
Human-adjudicated in fixed-claim/same-document audit. &
Source-controlled hard \textsc{NEI}. \\
Fixed-claim hard \textsc{NEI} &
Pair the same claim with reference evidence and validated insufficient evidence. &
Evidence-side differences may still contain shallow cues. &
Human-adjudicated in fixed-claim/same-document audit. &
Evidence-subst. diagnostic. \\
Missing-hop control &
Remove a required supporting fact from multi-hop evidence. &
Partial evidence over-interpreted as sufficient. &
Candidate-only in HoVer. &
External multi-hop control. \\
FEVER subset control &
Construct bounded Wikipedia verification controls with alternative \textsc{NEI} evidence. &
Dataset-specific lexical or retrieval artifacts. &
Subset control, not human-validated hard \textsc{NEI}. &
External shortcut control. \\
\bottomrule
\end{tabular}
\end{adjustbox}
\caption{Full NEI-CAP construction taxonomy. The taxonomy separates shortcut-prone baselines, candidate hard \textsc{NEI}, human-adjudicated hard \textsc{NEI}, and bounded external controls.}
\label{tab:app-full-taxonomy}
\end{table}

\subsection{Construction Strength}

We use three construction-strength categories.

\paragraph{Easy \textsc{NEI}.}
Placeholder, random irrelevant, and position-biased examples are shortcut-prone by design. They are useful probes of artifact sensitivity but should not be treated as strong evidence-sufficiency tests.

\paragraph{Candidate hard \textsc{NEI}.}
Candidate hard examples are constructed to be semantically related but insufficient. Before adjudication, they may contain semantic contamination.

\paragraph{Human-adjudicated hard \textsc{NEI}.}
Human-adjudicated hard examples are candidate hard examples labeled as truly insufficient rather than actually supportive, actually refuting, ambiguous, or invalid. These subsets support the strongest hard-\textsc{NEI} claims in the paper.

\subsection{Terminology and Claim Boundaries}

\begin{table}[H]
\centering
\scriptsize
\begin{adjustbox}{max width=\columnwidth}
\begin{tabular}{p{0.30\linewidth}p{0.32\linewidth}p{0.28\linewidth}}
\toprule
\textbf{Avoid} & \textbf{Use instead} & \textbf{Reason} \\
\midrule
hard-negative & hard \textsc{NEI} example & Avoids retrieval/contrastive ambiguity. \\
gold-truth & gold label / reference label & Standard terminology. \\
gold-side & reference-evidence side & More precise. \\
hard-side & insufficient-evidence side & Defines evidence condition. \\
counterfactual proof & fixed-claim evidence-substitution diagnostic & Avoids causal overclaim. \\
HoVer human-validated & HoVer candidate missing-hop control & No HoVer human audit. \\
FEVER full validation & FEVER subset control & Bounded external probe. \\
\bottomrule
\end{tabular}
\end{adjustbox}
\caption{Terminology mapping used throughout the paper.}
\label{tab:app-terminology}
\end{table}

The taxonomy supports three bounded claims: \textsc{NEI} evaluation is construction-sensitive; easy \textsc{NEI} can inflate apparent competence; and hard insufficient evidence should be validated when it supports central claims.

%% file: appendices/B_Dataset_Construction_and_Manifests.tex
\section{Dataset Construction and Manifests}
\label{app:data}

This appendix documents the dataset assets and manifest requirements behind NEI-CAP. The main paper reports only compact descriptions; the appendix records the construction metadata needed for reproducible construction-aware evaluation.

\subsection{Example Representation}

Each example is represented as:
\[
x_i=(c_i,E_i,y_i,z_i,g_i,m_i),
\]
where $c_i$ is the claim, $E_i$ is the evidence set, $y_i \in \{\textsc{Support},\textsc{Refute},\textsc{NEI}\}$ is the label, $z_i$ is the construction family, $g_i$ is a group identifier, and $m_i$ stores provenance and audit metadata. The construction variable $z_i$ is not given to the model; it is used for splitting, auditing, and reporting.

\subsection{Manifest Schema}

\begin{table}[H]
\centering
\scriptsize
\begin{adjustbox}{max width=\columnwidth}
\begin{tabular}{p{0.22\linewidth}p{0.18\linewidth}p{0.48\linewidth}}
\toprule
\textbf{Field} & \textbf{Type} & \textbf{Description} \\
\midrule
example ID & string & Unique claim--evidence instance ID. \\
claim ID & string & Original claim ID. \\
group ID & string & Group key for group-disjoint splitting. \\
source data & string & SciFact, FEVER, HoVer, or derived resource. \\
claim & string & Claim text. \\
evidence & string/list & Evidence text or list of evidence units. \\
label & categorical & \textsc{Support}, \textsc{Refute}, or \textsc{NEI}. \\
construction & categorical & Placeholder, BM25 near-miss, same-document, etc. \\
split & categorical & Train, development, test, or audit. \\
document ID & string/list & Source document IDs when available. \\
sentence IDs & string/list & Sentence or rationale IDs when available. \\
retrieval method & string & Retrieval or sampling method. \\
retrieval rank & integer & Rank of retrieved evidence. \\
BM25 score & float & Retrieval score, when applicable. \\
sentence position & integer/list & Sentence position within document. \\
validation status & categorical & Not validated, candidate, valid \textsc{NEI}, contaminated, or ambiguous. \\
adjudicated label & categorical & Final adjudicated label if available. \\
\bottomrule
\end{tabular}
\end{adjustbox}
\caption{Recommended NEI-CAP manifest schema. Construction metadata is required to report results by evidence condition rather than by label alone.}
\label{tab:app-manifest-schema}
\end{table}

\subsection{SciFact Suite}

The core SciFact suite contains five variants: placeholder, random irrelevant, position-biased, BM25 near-miss, and cited non-rationale. The \textsc{Support} and \textsc{Refute} portions remain comparable across variants; only the \textsc{NEI} evidence condition changes. This design supports train/test construction-shift evaluation.

\subsection{Group-Disjoint Splitting}

Construction variants derived from the same claim or claim--document group must not leak across train and test. NEI-CAP therefore uses group-disjoint splitting keyed by claim, claim--document pair, or fixed-claim substitution group where applicable. This prevents a model from exploiting claim memorization across evidence variants.

\subsection{Human-Adjudicated Assets}

Human-adjudicated assets are separated from automatically constructed candidate data. The SciFact audit validates BM25/cited candidate hard \textsc{NEI}; the fixed-claim/same-document audit validates same-claim and same-document diagnostics. These assets estimate semantic contamination and define human-adjudicated hard-\textsc{NEI} subsets for model evaluation.

\subsection{Human-Hard Evaluation Sampling Trail}

The SciFact human-audit pool estimates label validity, while the model evaluation uses only the held-out group-disjoint test split. Table~\ref{tab:app-human-hard-sampling-trail} records the paper-facing sampling trail used to interpret the $n=54$ hard-\textsc{NEI} evaluation in Table~\ref{tab:human-hard-nei-results}.

\begin{table}[H]
\centering
\scriptsize
\begin{adjustbox}{max width=\columnwidth}
\begin{tabular}{lr}
\toprule
\textbf{Stage} & \textbf{Count} \\
\midrule
Candidate BM25/cited hard-\textsc{NEI} audit pool & 250 \\
Human-valid \textsc{NEI} after adjudication & 223 \\
Human-adjudicated hard-\textsc{NEI} subtype & 195 \\
Held-out group-disjoint model-evaluation split & 54 \\
\bottomrule
\end{tabular}
\end{adjustbox}
\caption{Sampling trail for the SciFact human-hard model-evaluation subset. The larger audit pool estimates label validity; the held-out split supports model evaluation.}
\label{tab:app-human-hard-sampling-trail}
\end{table}

\subsection{External Controls}

FEVER and HoVer are included as bounded external controls. FEVER provides a non-toy Wikipedia subset control. HoVer provides a candidate missing-hop control for multi-hop insufficiency. Neither is treated as human-validated hard \textsc{NEI}.

\subsection{Construction Split Statistics and Leakage Audit}

Construction-level split statistics are recorded for each paper-facing family: label counts, claim and document group counts, evidence length, sentence count, placeholder rate, claim--evidence overlap, coverage, retrieval metadata, duplicate counts, and missing-field checks. Table~\ref{tab:app-additional-split-summary} gives a compact test-split view for the core SciFact suite; full CSV artifacts will be released with the accompanying code and data package.

\begin{table}[H]
\centering
\scriptsize
\begin{adjustbox}{max width=\columnwidth}
\begin{tabular}{lrrrr}
\toprule
\textbf{Variant} & \textbf{$n$} & \textbf{SUP/REF/NEI} & \textbf{Avg. tok.} & \textbf{Coverage} \\
\midrule
Placeholder & 177 & 76/40/61 & 174.7 & 0.444 \\
Random irrelevant & 177 & 76/40/61 & 236.8 & 0.474 \\
Position-biased & 177 & 76/40/61 & 181.7 & 0.501 \\
BM25 near-miss & 177 & 76/40/61 & 251.4 & 0.630 \\
Cited non-rationale & 177 & 76/40/61 & 246.9 & 0.590 \\
\bottomrule
\end{tabular}
\end{adjustbox}
\caption{Compact test-split statistics for the SciFact construction suite. Each variant keeps comparable \textsc{Support}/\textsc{Refute}/\textsc{NEI} label counts while changing the \textsc{NEI} evidence condition.}
\label{tab:app-additional-split-summary}
\end{table}

The leakage audit reports zero claim-group overlap across train/development/test for the core SciFact construction variants and zero construction-variant cross-split leakage. Document overlap can occur because the same scientific paper may be relevant to multiple claims; we report it as source-distribution metadata rather than claim leakage.

\subsection{Asset Map}

A future public release will link each reported result to its source dataset, construction family, split policy, model configuration, seed set, prediction artifact, and evaluation output. The full machine-readable asset map will be provided with the released manifests.

%% file: appendices/C_Artifact_Audit_Statistics.tex
\section{Artifact Audit Statistics}
\label{app:audit}

NEI-CAP audits construction families before interpreting model performance. The audit asks whether \textsc{NEI} examples can be separated using superficial evidence-side features rather than evidence-sufficiency reasoning.

\subsection{Audited Features}

We audit evidence length, number of evidence sentences, claim--evidence lexical overlap, claim--evidence coverage, placeholder rate, sentence position, source concentration, and discourse markers. These features do not determine semantic validity; they identify shortcut risk.

\subsection{Construction-Level Summary}

\begin{table}[H]
\centering
\scriptsize
\begin{adjustbox}{max width=\columnwidth}
\begin{tabular}{p{0.24\linewidth}rrrrp{0.16\linewidth}}
\toprule
\textbf{Construction group} & \textbf{$n$} & \textbf{Avg. sent.} & \textbf{Avg. tok.} & \textbf{Coverage} & \textbf{Status} \\
\midrule
SciFact BM25/cited hard \textsc{NEI} & 195 & 11.49 & 216.56 & 0.483 & Human-adjudicated \\
Same-document hard \textsc{NEI} & 114 & 1.04 & 23.93 & 0.233 & Human-adjudicated \\
Fixed-claim hard \textsc{NEI} & 114 & 1.23 & 23.93 & 0.233 & Human-adjudicated \\
Placeholder \textsc{NEI} & 61 & 1.00 & 2.00 & 0.005 & Constructed only \\
Random irrelevant \textsc{NEI} & 61 & 9.70 & 182.46 & 0.092 & Constructed only \\
\bottomrule
\end{tabular}
\end{adjustbox}
\caption{Artifact audit summary by construction group. Placeholder \textsc{NEI} has extremely short evidence and near-zero coverage; human-adjudicated hard \textsc{NEI} contains substantive evidence.}
\label{tab:app-artifact-summary}
\end{table}

\subsection{Overlap and Marker Statistics}

\begin{table}[H]
\centering
\scriptsize
\begin{adjustbox}{max width=\columnwidth}
\begin{tabular}{p{0.24\linewidth}rrrr}
\toprule
\textbf{Construction group} & \textbf{Overlap count} & \textbf{Jaccard} & \textbf{Context marker} & \textbf{Method marker} \\
\midrule
SciFact BM25/cited hard \textsc{NEI} & 6.24 & 0.053 & 0.210 & 0.226 \\
Same-document hard \textsc{NEI} & 3.06 & 0.098 & 0.316 & 0.140 \\
Fixed-claim hard \textsc{NEI} & 3.06 & 0.098 & 0.316 & 0.140 \\
Placeholder \textsc{NEI} & 0.05 & 0.005 & 0.000 & 0.000 \\
Random irrelevant \textsc{NEI} & 1.31 & 0.013 & 0.148 & 0.213 \\
\bottomrule
\end{tabular}
\end{adjustbox}
\caption{Claim--evidence overlap and marker statistics. Random irrelevant \textsc{NEI} contains long evidence but low overlap, indicating a topic-mismatch shortcut rather than evidence insufficiency.}
\label{tab:app-overlap-summary}
\end{table}

\subsection{Hard-NEI Subtypes}

\begin{table}[H]
\centering
\scriptsize
\begin{adjustbox}{max width=\columnwidth}
\begin{tabular}{lrrr}
\toprule
\textbf{Group} & \textbf{Broad topic} & \textbf{Near-miss} & \textbf{Partial} \\
\midrule
SciFact BM25/cited hard \textsc{NEI} & 62 & 102 & 31 \\
Fixed-claim/same-document hard \textsc{NEI} & 50 & 49 & 15 \\
\bottomrule
\end{tabular}
\end{adjustbox}
\caption{Subtype distribution for human-adjudicated hard \textsc{NEI}. Hard \textsc{NEI} is heterogeneous rather than a single condition.}
\label{tab:app-hard-nei-subtypes}
\end{table}

\subsection{Interpretation}

The audit supports three conclusions. First, placeholder \textsc{NEI} exposes strong format and absence cues. Second, random irrelevant \textsc{NEI} mainly tests topic mismatch. Third, human-adjudicated hard \textsc{NEI} is heterogeneous across BM25/cited, same-document, and fixed-claim settings. These findings motivate construction-specific reporting rather than aggregate \textsc{NEI}-F1 alone.

%% file: appendices/D_Human_Validation_Protocol_and_Adjudication.tex
\section{Human Validation Protocol and Adjudication}
\label{app:human-validation}

Hard insufficient evidence can be noisy. Retrieved near-misses, cited non-rationales, and same-document non-rationales may contain implicit support, implicit refutation, or ambiguity. NEI-CAP therefore separates automatically constructed candidate hard \textsc{NEI} from human-adjudicated hard \textsc{NEI}.

\subsection{Validation Goal}

Human validation determines whether a candidate hard-\textsc{NEI} example is truly insufficient. The resulting labels are used to estimate semantic contamination and to define human-adjudicated hard-\textsc{NEI} subsets for evaluation.

\subsection{Annotation Labels}

\begin{table}[H]
\centering
\scriptsize
\begin{adjustbox}{max width=\columnwidth}
\begin{tabular}{p{0.35\linewidth}p{0.55\linewidth}}
\toprule
\textbf{Label} & \textbf{Definition} \\
\midrule
\texttt{truly\_insufficient} & Evidence does not support or refute the claim. \\
\texttt{actually\_supported} & Evidence supports the claim. \\
\texttt{actually\_contradicted} & Evidence refutes the claim; reported as \textsc{Refute} in paper-facing terminology. \\
\texttt{ambiguous} & Status cannot be determined reliably. \\
\texttt{invalid\_or\_unreadable} & Claim or evidence is malformed, missing, unreadable, or out of scope. \\
\bottomrule
\end{tabular}
\end{adjustbox}
\caption{Human validation label schema. Contamination includes actually supported, actually contradicted/refuted, ambiguous, invalid, and unreadable cases.}
\label{tab:app-human-labels}
\end{table}

\subsection{Reporting Metrics}

We report:
\[
\mathrm{valid\ NEI}=\frac{N_{\mathrm{insufficient}}}{N_{\mathrm{audited}}},
\]
and:
\[
\mathrm{contam.}=\frac{N_{\mathrm{sup}}+N_{\mathrm{ref}}+N_{\mathrm{amb/invalid}}}{N_{\mathrm{audited}}}.
\]
AI-assisted checks, where used, are treated only as triage or secondary support, not as human validation.

\subsection{SciFact Hard-NEI Audit}

\begin{table}[H]
\centering
\scriptsize
\begin{adjustbox}{max width=\columnwidth}
\begin{tabular}{lrrr}
\toprule
\textbf{Metric} & \textbf{Observed} & \textbf{95\% low} & \textbf{95\% high} \\
\midrule
Valid \textsc{NEI} rate & 0.892 & 0.852 & 0.928 \\
Contamination rate & 0.108 & 0.072 & 0.148 \\
Actually supported rate & 0.056 & 0.028 & 0.088 \\
Actually refuted rate & 0.052 & 0.028 & 0.084 \\
Ambiguous/invalid rate & 0.000 & 0.000 & 0.000 \\
Hard subtype rate among valid \textsc{NEI} & 0.874 & 0.826 & 0.919 \\
Topic-unrelated rate among valid \textsc{NEI} & 0.117 & 0.076 & 0.164 \\
\bottomrule
\end{tabular}
\end{adjustbox}
\caption{SciFact hard-\textsc{NEI} audit validation over 250 candidate hard-\textsc{NEI} examples. Intervals are bootstrap 95\% intervals.}
\label{tab:app-exp008-validation}
\end{table}

\subsection{Fixed-Claim and Same-Document Validation}

\begin{table}[H]
\centering
\scriptsize
\begin{adjustbox}{max width=\columnwidth}
\begin{tabular}{lrrrr}
\toprule
\textbf{Subset} & \textbf{Rows} & \textbf{Contam.} & \textbf{Hard rows} & \textbf{Hard rate} \\
\midrule
Final adjudicated labels & 122 & 0.0328 & 114 & 0.966 \\
Human-validated \textsc{NEI} & 118 & 0.0000 & 114 & 0.966 \\
Human-validated hard \textsc{NEI} & 114 & 0.0000 & 114 & 1.000 \\
Same-claim hard \textsc{NEI} & 114 & 0.0000 & 114 & 1.000 \\
Same-document hard \textsc{NEI} & 114 & 0.0000 & 114 & 1.000 \\
\bottomrule
\end{tabular}
\end{adjustbox}
\caption{Fixed-claim/same-document validation summary. The 114 hard-\textsc{NEI} examples support same-claim and same-document diagnostics.}
\label{tab:app-fixedclaim-validation}
\end{table}

\subsection{Annotator and Adjudication Protocol}

Two annotators participated in human validation, and final labels are consensus adjudications. Annotators independently evaluated claim plus candidate evidence for semantic sufficiency. Blinded packets removed LLM labels, gold labels, model predictions, and construction provenance from primary annotation fields. AI-assisted checks were used only as secondary support or triage and are never reported as standalone human validation.

\subsection{Inter-Annotator Agreement}

Table~\ref{tab:app-iaa} reports formal agreement on the 250-example SciFact audit, computed from the two independent pre-consensus annotation streams with paired-row bootstrap intervals (10{,}000 resamples). Each decision is reported separately rather than pooled into a single coefficient~\citep{artstein2008inter}.

\begin{table}[H]
\centering
\scriptsize
\begin{adjustbox}{max width=\columnwidth}
\begin{tabular}{lrrrr}
\toprule
\textbf{Agreement metric} & \textbf{$n$} & \textbf{Raw} & \textbf{Cohen's $\kappa$} & \textbf{95\% CI} \\
\midrule
Five-way primary label & 250 & 94.4\% & 0.732 & [0.585, 0.854] \\
Valid \textsc{NEI} vs.\ contaminated & 250 & 94.8\% & 0.743 & [0.589, 0.866] \\
Subtype (cond., both \textsc{NEI}) & 215 & 54.9\% & 0.363 & [0.268, 0.454] \\
Hard vs.\ non-hard (cond., both \textsc{NEI}) & 215 & 93.0\% & 0.714 & [0.563, 0.837] \\
\bottomrule
\end{tabular}
\end{adjustbox}
\caption{Inter-annotator agreement on the SciFact hard-\textsc{NEI} audit. The conditional rows are computed only on the subset where both annotators independently assigned \textsc{NEI}; they are never pooled with the $n=250$ figures.}
\label{tab:app-iaa}
\end{table}

The decisions the paper relies on are stable across annotators: five-way primary-label agreement is 236/250 (94.4\%), and agreement on the binary valid-\textsc{NEI}-versus-contamination gate that defines the hard subsets is 94.8\%. Conditional on both annotators assigning \textsc{NEI} ($n=215$), exact insufficiency-subtype agreement is lower (54.9\%, $\kappa=0.363$), while hard-versus-non-hard agreement remains high (93.0\%, $\kappa=0.714$). Of the 97 subtype disagreements, 82 remain within the same hard/non-hard category; only 15 cross the evaluation gate. Fine-grained subtype boundaries are therefore comparatively subjective, which is consistent with treating hard \textsc{NEI} as heterogeneous (Section~\ref{subsec:hard-nei-heterogeneity}), while the coarser decisions remain reliable. The 14 primary-label disagreements were resolved through joint adjudication; the resulting consensus labels are the labels used throughout the paper.

\subsection{Boundary}

The paper may claim that the SciFact hard-\textsc{NEI} audit and the fixed-claim/same-document audit provide human-adjudicated hard-\textsc{NEI} subsets. It should not claim that human validation eliminates all ambiguity, that candidate-only examples are valid without adjudication, or that FEVER/HoVer controls are human-validated hard \textsc{NEI}.

%% file: appendices/E_Experimental_Setup_and_Model_Details.tex
\section{Experimental Setup and Model Details}
\label{app:setup}

NEI-CAP uses models as diagnostic probes of construction sensitivity, not as proposed architectures.

\subsection{Task Format}

All neural experiments are three-way claim--evidence classification:
\[
y \in \{\textsc{Support},\textsc{Refute},\textsc{NEI}\}.
\]
The model receives claim and evidence text. Construction family is not provided as model input; it defines train/evaluation variants and reporting groups.

\subsection{Primary SciFact Configuration}

\begin{table}[H]
\centering
\scriptsize
\begin{adjustbox}{max width=\columnwidth}
\begin{tabular}{p{0.38\linewidth}p{0.50\linewidth}}
\toprule
\textbf{Setting} & \textbf{Value} \\
\midrule
Primary model & \texttt{deberta-v3-base} \\
Input root & NEI-CAP SciFact suite \\
Train variants & placeholder, cited, random irrelevant, BM25 near-miss, position-biased \\
Evaluation variants & placeholder, cited, random irrelevant, BM25 near-miss, position-biased \\
Evaluation split & test \\
Max length & 384 \\
Epochs & 3 \\
Batch size & 16 \\
Learning rate & $2.0\times10^{-5}$ \\
Weight decay & 0.01 \\
Warmup ratio & 0.10 \\
Class weighting & enabled \\
Early stopping metric & development Macro-F1 \\
Patience & 1 \\
Precision & bf16 if available, otherwise fp32 \\
Seeds & 13, 17, 23 \\
Checkpoint policy & save best checkpoint \\
Prediction logging & enabled \\
\bottomrule
\end{tabular}
\end{adjustbox}
\caption{Primary SciFact construction-matrix configuration.}
\label{tab:app-exp009-config}
\end{table}

\subsection{Construction Matrix}

The primary matrix trains one verifier for each \textsc{NEI} construction and evaluates it on all five constructions. Each cell asks whether performance transfers across evidence conditions while \textsc{Support} and \textsc{Refute} portions remain comparable.

\subsection{Secondary Backbones}

RoBERTa-base and SciBERT are secondary robustness probes. They reuse the same construction suite and train/evaluation variants. Secondary models supplement, but do not replace, the primary DeBERTa matrix.

\subsection{Metrics}

For three-way verification, we report accuracy, Macro-F1, class-specific F1, and especially \textsc{NEI}-F1. For one-class human-hard evaluation, Macro-F1 is not meaningful. We report \textsc{NEI} recall, false \textsc{Support} rate, false \textsc{Refute} rate, and mean predicted class probabilities.

\subsection{Robustness Classification}

For secondary-backbone SciFact probes, construction sensitivity is considered replicated when placeholder matched \textsc{NEI}-F1 is high, placeholder-to-hard drop is large, and at least one hard construction has non-trivial matched \textsc{NEI}-F1. These thresholds are diagnostic for this study, not universal standards.

\subsection{Five-Seed and Mixed-Construction Configuration}

The five-seed robustness and mixed-construction experiments use seeds \{13,17,23,29,37\}. Mixed-construction regimes include an easy mixture over shortcut-prone constructions, a hard mixture over BM25/cited constructions, and a balanced mixture over all five SciFact construction families. These experiments test whether construction sensitivity persists under more realistic training distributions.

\subsection{Result Provenance}

Each reported result should be traceable to a dataset manifest, construction family, model configuration, seed set, prediction file, evaluation output, and paper table or figure. Aggregate scores without construction metadata are insufficient for NEI-CAP.

%% file: appendices/F_Full_SciFact_Construction_Matrices.tex
\section{Full SciFact Construction Matrices}
\label{app:scifact-matrices}

This appendix reports the full DeBERTa SciFact construction-shift matrices.

\subsection{NEI-F1 Matrix}

\begin{table}[H]
\centering
\scriptsize
\begin{adjustbox}{max width=\columnwidth}
\begin{tabular}{lrrrrr}
\toprule
\textbf{Train NEI} & \textbf{BM25} & \textbf{Cited} & \textbf{Placeholder} & \textbf{Position} & \textbf{Random} \\
\midrule
BM25 near-miss & 0.675 & 0.561 & 0.725 & 0.652 & 0.676 \\
Cited non-rationale & 0.508 & 0.578 & 0.768 & 0.768 & 0.599 \\
Placeholder & 0.000 & 0.000 & 1.000 & 0.947 & 0.000 \\
Position-biased & 0.000 & 0.000 & 1.000 & 1.000 & 0.063 \\
Random irrelevant & 0.445 & 0.294 & 0.995 & 0.992 & 0.995 \\
\bottomrule
\end{tabular}
\end{adjustbox}
\caption{Full SciFact construction-shift matrix for \textsc{NEI}-F1. Rows are train constructions; columns are evaluation constructions.}
\label{tab:app-nei-f1-matrix}
\end{table}

\subsection{Macro-F1 Matrix}

\begin{table}[H]
\centering
\scriptsize
\begin{adjustbox}{max width=\columnwidth}
\begin{tabular}{lrrrrr}
\toprule
\textbf{Train NEI} & \textbf{BM25} & \textbf{Cited} & \textbf{Placeholder} & \textbf{Position} & \textbf{Random} \\
\midrule
BM25 near-miss & 0.529 & 0.479 & 0.559 & 0.530 & 0.532 \\
Cited non-rationale & 0.357 & 0.386 & 0.474 & 0.474 & 0.399 \\
Placeholder & 0.277 & 0.277 & 0.684 & 0.657 & 0.276 \\
Position-biased & 0.262 & 0.262 & 0.665 & 0.665 & 0.285 \\
Random irrelevant & 0.474 & 0.415 & 0.716 & 0.715 & 0.716 \\
\bottomrule
\end{tabular}
\end{adjustbox}
\caption{Full SciFact construction-shift matrix for Macro-F1. Aggregate performance can appear strong under easy matched evaluation while hiding hard-\textsc{NEI} failure.}
\label{tab:app-macro-f1-matrix}
\end{table}

\subsection{Drop Summary}

BM25 and cited training do not yield perfect matched scores, but they transfer more stably across hard constructions.

\begin{table}[H]
\centering
\scriptsize
\begin{adjustbox}{max width=\columnwidth}
\begin{tabular}{lrrrr}
\toprule
\textbf{Train NEI} & \textbf{Matched} & \textbf{BM25} & \textbf{Cited} & \textbf{Hard drop} \\
\midrule
Placeholder & 1.000 & 0.000 & 0.000 & 1.000 \\
Position-biased & 1.000 & 0.000 & 0.000 & 1.000 \\
Random irrelevant & 0.995 & 0.445 & 0.294 & 0.625 \\
BM25 near-miss & 0.675 & 0.675 & 0.561 & 0.057 \\
Cited non-rationale & 0.578 & 0.508 & 0.578 & 0.035 \\
\bottomrule
\end{tabular}
\end{adjustbox}
\caption{Matched and hard-evaluation \textsc{NEI}-F1 comparisons. Hard drop compares matched performance against average BM25/cited performance.}
\label{tab:app-drop-summary}
\end{table}

%% file: appendices/G_Human_Validated_Hard_NEI_Evaluation.tex
\section{Human-Validated Hard-NEI Evaluation}
\label{app:human-hard-eval}

This appendix reports full evaluation on the held-out SciFact human-adjudicated hard-\textsc{NEI} test split ($n=54$). The larger audit pool and sampling trail are documented in Appendix~\ref{app:data}; since all evaluated examples are validated as \textsc{NEI}, Macro-F1 is not reported.

\subsection{Recall and Error Rates}

\begin{table}[H]
\centering
\scriptsize
\begin{adjustbox}{max width=\columnwidth}
\begin{tabular}{lrrrrrr}
\toprule
\textbf{Train NEI} & \textbf{Recall} & \textbf{95\% low} & \textbf{95\% high} & \textbf{False SUP} & \textbf{False REF} & \textbf{$n$} \\
\midrule
BM25 near-miss & 0.691 & 0.593 & 0.790 & 0.142 & 0.167 & 54 \\
Cited non-rationale & 0.679 & 0.562 & 0.790 & 0.074 & 0.247 & 54 \\
Random irrelevant & 0.216 & 0.123 & 0.321 & 0.574 & 0.210 & 54 \\
Placeholder & 0.000 & 0.000 & 0.000 & 0.870 & 0.130 & 54 \\
Position-biased & 0.000 & 0.000 & 0.000 & 0.907 & 0.093 & 54 \\
\bottomrule
\end{tabular}
\end{adjustbox}
\caption{Primary DeBERTa evaluation on human-adjudicated hard \textsc{NEI}. False REF maps internal contradiction errors to the paper-facing \textsc{Refute} label.}
\label{tab:app-human-hard-recall}
\end{table}

Bootstrap intervals degenerate for the exact 0/54 outcomes: resampling a subset with zero successes always yields zero. For placeholder and position-biased training we therefore additionally report the Clopper--Pearson exact 95\% interval~\citep{clopper1934use}, $[0.000, 0.066]$. This interval does not overlap the BM25 near-miss bootstrap interval of $[0.593, 0.790]$.

\subsection{Predicted Probabilities}

\begin{table}[H]
\centering
\scriptsize
\begin{adjustbox}{max width=\columnwidth}
\begin{tabular}{lrrrrrr}
\toprule
\textbf{Train NEI} & \textbf{Mean $P$(\textsc{NEI})} & \textbf{95\% low} & \textbf{95\% high} & \textbf{Mean $P$(SUP)} & \textbf{Mean $P$(REF)} & \textbf{$n$} \\
\midrule
BM25 near-miss & 0.398 & 0.366 & 0.431 & 0.324 & 0.277 & 54 \\
Cited non-rationale & 0.416 & 0.371 & 0.458 & 0.318 & 0.266 & 54 \\
Random irrelevant & 0.212 & 0.129 & 0.304 & 0.417 & 0.371 & 54 \\
Placeholder & 0.013 & 0.012 & 0.013 & 0.540 & 0.448 & 54 \\
Position-biased & 0.014 & 0.013 & 0.014 & 0.551 & 0.435 & 54 \\
\bottomrule
\end{tabular}
\end{adjustbox}
\caption{Mean predicted probabilities on human-adjudicated hard \textsc{NEI}. Easy training assigns near-zero \textsc{NEI} probability to validated hard insufficient evidence and reallocates probability mass to \textsc{Support} and \textsc{Refute}.}
\label{tab:app-human-hard-prob}
\end{table}

%% file: appendices/H_Multi_Model_Robustness.tex
\section{Multi-Model Robustness}
\label{app:multimodel}

This appendix documents secondary-backbone robustness, mixed-construction training, and same-document diagnostics. RoBERTa and SciBERT are diagnostic probes that supplement the primary DeBERTa matrix.

\subsection{Secondary Backbones on Human-Adjudicated Hard NEI}

\begin{table}[H]
\centering
\scriptsize
\begin{adjustbox}{max width=\columnwidth}
\begin{tabular}{llrrrr}
\toprule
\textbf{Model} & \textbf{Train NEI} & \textbf{Recall} & \textbf{False SUP} & \textbf{False REF} & \textbf{Mean $P$(\textsc{NEI})} \\
\midrule
SciBERT & BM25 near-miss & 0.838 & 0.070 & 0.092 & 0.710 \\
SciBERT & Cited non-rationale & 0.875 & 0.063 & 0.062 & 0.692 \\
SciBERT & Placeholder & 0.000 & 0.574 & 0.426 & 0.011 \\
SciBERT & Position-biased & 0.007 & 0.838 & 0.156 & 0.024 \\
SciBERT & Random irrelevant & 0.099 & 0.520 & 0.381 & 0.114 \\
\midrule
RoBERTa & BM25 near-miss & 0.703 & 0.116 & 0.181 & 0.404 \\
RoBERTa & Cited non-rationale & 0.638 & 0.109 & 0.253 & 0.435 \\
RoBERTa & Placeholder & 0.000 & 0.867 & 0.133 & 0.004 \\
RoBERTa & Position-biased & 0.009 & 0.932 & 0.060 & 0.014 \\
RoBERTa & Random irrelevant & 0.186 & 0.586 & 0.227 & 0.191 \\
\bottomrule
\end{tabular}
\end{adjustbox}
\caption{Secondary-backbone evaluation on human-adjudicated hard \textsc{NEI}. Placeholder-trained RoBERTa and SciBERT both collapse on validated hard insufficient evidence.}
\label{tab:app-multimodel-human-hard}
\end{table}

\subsection{Mixed-Construction Training}

Table~\ref{tab:app-mixed-training-summary} summarizes DeBERTa mixed-construction training. Easy mixtures remain weak on hard conditions, while hard and balanced mixtures improve hard-\textsc{NEI} recognition.

\begin{table}[H]
\centering
\scriptsize
\begin{adjustbox}{max width=\columnwidth}
\begin{tabular}{lrrr}
\toprule
\textbf{Regime} & \textbf{BM25} & \textbf{Cited} & \textbf{SciFact hard} \\
\midrule
Easy mixture & 0.379 & 0.273 & 0.178 \\
Hard mixture & 0.662 & 0.652 & 0.770 \\
Balanced mixture & 0.802 & 0.803 & 0.915 \\
\bottomrule
\end{tabular}
\end{adjustbox}
\caption{DeBERTa mixed-construction training summary. BM25 and Cited are construction-stratified \textsc{NEI}-F1; SciFact hard is one-class \textsc{NEI} recall on human-adjudicated hard \textsc{NEI}. Same-document results are reported separately because that family is constructionally distinct.}
\label{tab:app-mixed-training-summary}
\end{table}

\begin{table}[H]
\centering
\scriptsize
\begin{adjustbox}{max width=\columnwidth}
\begin{tabular}{lrrrrr}
\toprule
\textbf{Regime} & \textbf{PH} & \textbf{Position} & \textbf{Random} & \textbf{BM25} & \textbf{Cited} \\
\midrule
Easy mixture & 0.989 & 0.989 & 0.960 & 0.379 & 0.273 \\
Hard mixture & 0.754 & 0.729 & 0.694 & 0.662 & 0.652 \\
Balanced mixture & 0.903 & 0.903 & 0.901 & 0.802 & 0.803 \\
\bottomrule
\end{tabular}
\end{adjustbox}
\caption{Full DeBERTa mixed-construction stratified \textsc{NEI}-F1 matrix. The balanced mixture improves all conditions but still yields a construction-specific profile: easy conditions remain higher than BM25/cited hard conditions. PH denotes placeholder.}
\label{tab:app-mixed-training-stratified}
\end{table}

\subsection{Same-Document Robustness}

\begin{table}[H]
\centering
\scriptsize
\begin{adjustbox}{max width=\columnwidth}
\begin{tabular}{llrrr}
\toprule
\textbf{Model} & \textbf{Train NEI} & \textbf{Recall} & \textbf{False SUP} & \textbf{False REF} \\
\midrule
SciBERT & BM25 near-miss & 0.482 & 0.231 & 0.287 \\
SciBERT & Cited non-rationale & 0.731 & 0.143 & 0.126 \\
SciBERT & Placeholder & 0.000 & 0.509 & 0.491 \\
SciBERT & Position-biased & 1.000 & 0.000 & 0.000 \\
SciBERT & Random irrelevant & 0.216 & 0.415 & 0.368 \\
\midrule
RoBERTa & BM25 near-miss & 0.816 & 0.023 & 0.161 \\
RoBERTa & Cited non-rationale & 0.871 & 0.000 & 0.129 \\
RoBERTa & Placeholder & 0.526 & 0.412 & 0.061 \\
RoBERTa & Position-biased & 1.000 & 0.000 & 0.000 \\
RoBERTa & Random irrelevant & 0.950 & 0.029 & 0.020 \\
\bottomrule
\end{tabular}
\end{adjustbox}
\caption{Secondary-backbone same-document evaluation. Same-document hard \textsc{NEI} behaves differently from SciFact BM25/cited hard \textsc{NEI}, reinforcing hard-\textsc{NEI} heterogeneity.}
\label{tab:app-multimodel-samedoc}
\end{table}

\subsection{Boundary}

The secondary models support the claim that construction sensitivity persists beyond DeBERTa. They do not show that all model families behave identically, and they do not replace the primary SciFact matrix.

%% file: appendices/I_Same_Claim_and_Same_Document_Diagnostics.tex
\section{Same-Claim and Same-Document Diagnostics}
\label{app:same-claim}

Fixed-claim/same-document diagnostics control the claim or source document while varying evidence. They are decomposition tests, not proofs of clean counterfactual evidence use.

\subsection{Setup}

The same-claim diagnostic pairs the same claim with reference evidence and with human-adjudicated insufficient evidence. The same-document diagnostic draws insufficient evidence from the same source document as the reference evidence. We report hard-\textsc{NEI} recall, false \textsc{Support}/\textsc{Refute} rates, reference-side accuracy, probability-drop success, strict swap success, and the mean drop in the reference-label probability. The reference label is \textsc{Support} or \textsc{Refute}, so this metric tracks non-\textsc{NEI} verification confidence rather than only \textsc{NEI} recall.

\subsection{Primary DeBERTa Fixed-Claim Diagnostics}

\begin{table}[H]
\centering
\scriptsize
\begin{adjustbox}{max width=\columnwidth}
\begin{tabular}{lrrr}
\toprule
\textbf{Train NEI} & \textbf{Hard recall} & \textbf{Swap success} & \textbf{Mean $\Delta$ ref} \\
\midrule
BM25 near-miss & 0.661 & 0.535 & 0.017 \\
Cited non-rationale & 0.988 & 0.743 & 0.023 \\
Placeholder & 0.898 & 0.737 & 0.112 \\
Position-biased & 1.000 & 0.728 & 0.034 \\
Random irrelevant & 0.947 & 0.807 & 0.124 \\
\bottomrule
\end{tabular}
\end{adjustbox}
\caption{Primary DeBERTa fixed-claim diagnostics. The same claim is paired with reference evidence and with insufficient evidence, so swap success and mean reference-label probability drop are defined.}
\label{tab:app-fixed-claim-primary}
\end{table}

\subsection{Primary DeBERTa Same-Document Hard-NEI Diagnostics}

\begin{table}[H]
\centering
\scriptsize
\begin{adjustbox}{max width=\columnwidth}
\begin{tabular}{lrrr}
\toprule
\textbf{Train NEI} & \textbf{Hard recall} & \textbf{False SUP} & \textbf{False REF} \\
\midrule
BM25 near-miss & 0.661 & 0.310 & 0.029 \\
Cited non-rationale & 0.988 & 0.009 & 0.003 \\
Placeholder & 0.898 & 0.085 & 0.018 \\
Position-biased & 1.000 & 0.000 & 0.000 \\
Random irrelevant & 0.947 & 0.053 & 0.000 \\
\bottomrule
\end{tabular}
\end{adjustbox}
\caption{Primary DeBERTa same-document hard-\textsc{NEI} diagnostics. The hard subset is shared with the fixed-claim diagnostic, so hard recall matches Table~\ref{tab:app-fixed-claim-primary}; same-document evaluation does not define swap success or reference-label probability drop, and instead reports error rates on the insufficient-evidence side.}
\label{tab:app-samedoc-primary}
\end{table}

\subsection{Secondary-Backbone Same-Claim Diagnostics}

\begin{table}[H]
\centering
\scriptsize
\begin{adjustbox}{max width=\columnwidth}
\begin{tabular}{llrrrr}
\toprule
\textbf{Model} & \textbf{Train NEI} & \textbf{Ref. acc.} & \textbf{Hard recall} & \textbf{Prob. drop} & \textbf{Strict swap} \\
\midrule
SciBERT & Placeholder & 0.599 & 0.000 & 0.494 & 0.000 \\
SciBERT & Cited non-rationale & 0.488 & 0.731 & 0.827 & 0.365 \\
SciBERT & Random irrelevant & 0.556 & 0.216 & 0.602 & 0.140 \\
SciBERT & BM25 near-miss & 0.529 & 0.482 & 0.734 & 0.289 \\
SciBERT & Position-biased & 0.006 & 1.000 & 0.883 & 0.006 \\
\midrule
RoBERTa & Placeholder & 0.401 & 0.503 & 0.643 & 0.137 \\
RoBERTa & Cited non-rationale & 0.108 & 0.833 & 0.614 & 0.053 \\
RoBERTa & Random irrelevant & 0.091 & 0.956 & 0.760 & 0.079 \\
RoBERTa & BM25 near-miss & 0.135 & 0.763 & 0.605 & 0.050 \\
RoBERTa & Position-biased & 0.009 & 1.000 & 0.538 & 0.009 \\
\bottomrule
\end{tabular}
\end{adjustbox}
\caption{Secondary-backbone same-claim diagnostics. Results motivate the bounded interpretation: hard-side recognition and reference-side verification can diverge.}
\label{tab:app-secondary-sameclaim}
\end{table}

\subsection{Same-Document Artifact Audit}

We audit whether the same-document hard-\textsc{NEI} family has a shallow construction signature. A shallow feature classifier nearly separates same-document hard \textsc{NEI} from several other construction groups using evidence length, sentence count, overlap, and coverage features. Table~\ref{tab:app-samedoc-artifact-audit} reports the compact audit result. The matched length/coverage audit is underpowered, so matched-subset comparisons are not conclusive.

\begin{table}[H]
\centering
\scriptsize
\begin{adjustbox}{max width=\columnwidth}
\begin{tabular}{lrrl}
\toprule
\textbf{Comparison} & \textbf{Acc.} & \textbf{Macro-F1} & \textbf{Status} \\
\midrule
Same-doc vs BM25/cited hard & 0.997 & 0.997 & completed \\
Same-doc vs placeholder & 1.000 & 1.000 & completed \\
Same-doc vs random irrelevant & 1.000 & 1.000 & completed \\
Same-doc vs constructed BM25/cited & 1.000 & 1.000 & completed \\
Matched feature audit & -- & -- & underpowered \\
\bottomrule
\end{tabular}
\end{adjustbox}
\caption{Same-document artifact audit. Same-document hard \textsc{NEI} is human-adjudicated but constructionally distinct; it should be reported as a source-controlled diagnostic family, not as artifact-free universal hard \textsc{NEI}.}
\label{tab:app-samedoc-artifact-audit}
\end{table}

\subsection{Interpretation}

The positive mean reference-label deltas in Table~\ref{tab:app-fixed-claim-primary} show that substituting insufficient evidence lowers confidence in the original \textsc{Support}/\textsc{Refute} label, and the secondary-backbone rows show that a model can succeed on the hard-\textsc{NEI} side while remaining weak on the reference-evidence side.

%% file: appendices/J_External_Controls_FEVER_and_HoVer.tex
\section{External Controls: FEVER and HoVer}
\label{app:external-controls}

FEVER and HoVer are bounded external controls: they test whether construction-aware diagnostics are useful beyond SciFact.

\subsection{FEVER Subset Control}

\begin{table}[H]
\centering
\scriptsize
\begin{adjustbox}{max width=\columnwidth}
\begin{tabular}{llrrrr}
\toprule
\textbf{Baseline} & \textbf{Train NEI} & \textbf{Placeholder} & \textbf{BM25} & \textbf{Random} & \textbf{PH$\rightarrow$BM25 drop} \\
\midrule
Claim+evidence TF-IDF & Placeholder & 1.000 & 0.000 & 0.000 & 1.000 \\
Claim+evidence TF-IDF & BM25 near-miss & 0.614 & 0.373 & 0.327 & 0.241 \\
Claim+evidence TF-IDF & Random irrelevant & 0.237 & 0.414 & 0.476 & -0.177 \\
Evidence-only TF-IDF & BM25 near-miss & 0.826 & 0.313 & 0.274 & 0.513 \\
Length/overlap logistic regression & BM25 near-miss & 0.772 & 0.406 & 0.536 & 0.366 \\
\bottomrule
\end{tabular}
\end{adjustbox}
\caption{FEVER external subset control. Placeholder-trained shallow models solve placeholder \textsc{NEI} while failing on BM25 near-miss and random-irrelevant evaluation.}
\label{tab:app-fever-control}
\end{table}

\subsection{HoVer Candidate Missing-Hop Control}

\begin{table}[H]
\centering
\scriptsize
\begin{adjustbox}{max width=\columnwidth}
\begin{tabular}{llrrrr}
\toprule
\textbf{Train} & \textbf{Eval} & \textbf{\textsc{NEI}-F1} & \textbf{\textsc{NEI} recall} & \textbf{False SUP} & \textbf{$n_{\mathrm{eval}}$} \\
\midrule
Missing-hop & Missing-hop & 0.835 & 0.756 & 0.097 & 7{,}978 \\
Missing-hop & Placeholder & 0.973 & 1.000 & 0.000 & 7{,}978 \\
Placeholder & Missing-hop & 0.000 & 0.000 & 0.431 & 7{,}978 \\
Placeholder & Placeholder & 1.000 & 1.000 & 0.000 & 7{,}978 \\
\bottomrule
\end{tabular}
\end{adjustbox}
\caption{HoVer candidate missing-hop control. Values are rounded to three decimals. HoVer rows are candidate-only and are not human-validated hard \textsc{NEI}.}
\label{tab:app-hover-control}
\end{table}

\subsection{Boundary}

FEVER supports a non-toy subset-level shortcut-control claim. HoVer supports a candidate-only multi-hop construction-sensitivity claim. Neither supports full cross-dataset human-validated hard-\textsc{NEI} generalization.

%% file: appendices/K_Diagnostic_Case_Studies.tex
\section{Diagnostic Case Studies}
\label{app:case-studies}

This appendix provides qualitative examples from row-level artifacts. We include only cases with available claim/evidence text and, where model behavior is discussed, saved predictions and probabilities. We do not reconstruct unavailable cases from aggregate statistics.

\subsection{Case Selection Policy}

Cases are selected only when: the claim and evidence are available, prediction and probability fields are available when model behavior is discussed, and human validation status is available when the case is described as human-adjudicated hard \textsc{NEI}. FEVER and HoVer cases are construction examples or candidate controls unless human-adjudicated.

\subsection{Fixed-Claim Hard-Side NEI Recognition}

\begin{table}[H]
\centering
\scriptsize
\begin{adjustbox}{max width=\columnwidth}
\begin{tabular}{p{0.20\linewidth}p{0.70\linewidth}}
\toprule
\textbf{Field} & \textbf{Value} \\
\midrule
Dataset & SciFact fixed-claim/same-document \\
Claim & Cholesterol loading induces KLF4 expression in vascular smooth muscle cells, resulting in pro-inflammatory cytokine expression. \\
Construction & Same-claim / same-document human-validated hard \textsc{NEI} \\
Train/seed & Random irrelevant / 17 \\
Ref. pred. & \textsc{Support} \\
Hard pred. & \textsc{NEI} \\
Prob. & $P(\mathrm{reference})=0.594$; $P(\mathrm{NEI}\mid E_{\mathrm{hard}})=0.978$ \\
Why it matters & The model changes prediction when evidence changes while the claim is fixed, supporting hard-side insufficiency recognition. \\
\bottomrule
\end{tabular}
\end{adjustbox}
\caption{Case study: fixed-claim hard-side \textsc{NEI} recognition.}
\label{tab:app-case-hard-side}
\end{table}

\subsection{Reference-Side Weakness}

\begin{table}[H]
\centering
\scriptsize
\begin{adjustbox}{max width=\columnwidth}
\begin{tabular}{p{0.20\linewidth}p{0.70\linewidth}}
\toprule
\textbf{Field} & \textbf{Value} \\
\midrule
Dataset & SciFact fixed-claim/same-document \\
Claim & T cell receptor/CD3 microdomains are required to induce the immunologic synapse. \\
Construction & Same-claim reference-side weakness \\
Train/seed & Position-biased / 13 \\
Ref. pred. & \textsc{NEI} \\
Hard pred. & \textsc{NEI} \\
Prob. & $P(\mathrm{reference})=0.001$; $P(\mathrm{NEI}\mid E_{\mathrm{hard}})=0.998$ \\
Why it matters & The hard side is recognized as \textsc{NEI}, but the reference-evidence side is missed, motivating the same-claim caveat. \\
\bottomrule
\end{tabular}
\end{adjustbox}
\caption{Case study: fixed-claim reference-side weakness.}
\label{tab:app-case-reference-weak}
\end{table}

\subsection{Same-Document Human-Validated Hard NEI}

\begin{table}[H]
\centering
\scriptsize
\begin{adjustbox}{max width=\columnwidth}
\begin{tabular}{p{0.20\linewidth}p{0.70\linewidth}}
\toprule
\textbf{Field} & \textbf{Value} \\
\midrule
Dataset & SciFact fixed-claim/same-document \\
Claim & Removal of H3K9me3 improves reprogramming efficiency in human somatic cell nuclear transfer experiments. \\
Evidence excerpt & Aberrant epigenetic reprogramming can cause developmental defects in somatic cell nuclear transfer embryos. \\
Status & Human-validated hard \textsc{NEI} \\
Prediction & \textsc{NEI} \\
Prob. & $P(\mathrm{NEI})=0.998$; $P(\mathrm{Support})=0.001$; $P(\mathrm{Refute})=0.001$ \\
Why it matters & Same-document evidence controls source/topic better than random irrelevant evidence while remaining insufficient. \\
\bottomrule
\end{tabular}
\end{adjustbox}
\caption{Case study: same-document human-validated hard \textsc{NEI}.}
\label{tab:app-case-samedoc}
\end{table}

\subsection{HoVer Candidate Missing-Hop Control}

\begin{table}[H]
\centering
\scriptsize
\begin{adjustbox}{max width=\columnwidth}
\begin{tabular}{p{0.20\linewidth}p{0.70\linewidth}}
\toprule
\textbf{Field} & \textbf{Value} \\
\midrule
Dataset & HoVer \\
Claim & Brett Herron's team competes in the Pro14 and the European Rugby Champions Cup. \\
Construction & Missing-one-supporting-fact candidate control \\
Status & Candidate-only, not human-validated \\
Train/seed & HoVer placeholder / 13 \\
Prediction & \textsc{Support} \\
Prob. & $P(\mathrm{Support})=0.981$; $P(\mathrm{NEI})=0.000$; $P(\mathrm{Refute})=0.018$ \\
Why it matters & Placeholder-trained behavior can fail on topically related missing-hop evidence. \\
\bottomrule
\end{tabular}
\end{adjustbox}
\caption{Case study: HoVer candidate missing-hop control.}
\label{tab:app-case-hover}
\end{table}

\subsection{Unavailable Row-Level Cases}

\begin{table}[H]
\centering
\scriptsize
\begin{adjustbox}{max width=\columnwidth}
\begin{tabular}{p{0.34\linewidth}p{0.54\linewidth}}
\toprule
\textbf{Case type} & \textbf{Unavailable reason} \\
\midrule
SciFact hard-audit placeholder failure on human-validated BM25/cited hard \textsc{NEI} & Row-level model predictions for final hard-\textsc{NEI} examples are unavailable; aggregate bootstrap tables exist only. \\
SciFact BM25/cited recovery case & Row-level paired recovery predictions for placeholder versus BM25/cited hard \textsc{NEI} are unavailable in primary outputs. \\
\bottomrule
\end{tabular}
\end{adjustbox}
\caption{Unavailable case-study reasons. We do not invent row-level cases when prediction artifacts are missing.}
\label{tab:app-unavailable-cases}
\end{table}

%% file: appendices/L_Statistical_Testing_and_Uncertainty_Estimation.tex
\section{Statistical Testing and Uncertainty Estimation}
\label{app:statistics}

This appendix defines metrics and uncertainty procedures for construction matrices, one-class hard-\textsc{NEI} evaluation, fixed-claim diagnostics, and prediction coverage checks.

\subsection{Three-Way Verification Metrics}

For standard verification, models predict:
\[
y \in \{\textsc{Support},\textsc{Refute},\textsc{NEI}\}.
\]
We report accuracy, Macro-F1, class-specific F1, and especially \textsc{NEI}-F1. Construction-shift matrices emphasize \textsc{NEI}-F1 because the experimental manipulation changes the \textsc{NEI} evidence condition.

\subsection{One-Class Human-Hard Metrics}

For human-validated hard-\textsc{NEI} subsets, all evaluated examples are adjudicated as \textsc{NEI}. Macro-F1 is therefore not informative. We report:
\[
\mathrm{\textsc{NEI}\ recall}=\frac{\#\text{predicted NEI}}{\#\text{validated NEI}},
\]
\[
\begin{aligned}
\mathrm{False\ Support} &= \frac{N_{\mathrm{pred=SUP}}}{N_{\mathrm{validated\ NEI}}},\\
\mathrm{False\ Refute} &= \frac{N_{\mathrm{pred=REF}}}{N_{\mathrm{validated\ NEI}}}.
\end{aligned}
\]
We also report mean predicted probabilities for \textsc{NEI}, \textsc{Support}, and \textsc{Refute}.

\subsection{Seed Aggregation and Bootstrap Intervals}

The primary construction matrix uses seeds 13, 17, and 23. The five-seed robustness and mixed-construction experiments use seeds 13, 17, 23, 29, and 37. Where seed-level outputs are available, we report mean, standard deviation, minimum, maximum, and bounded intervals. Human validation and human-hard evaluations report bootstrap 95\% intervals where available, computed by resampling evaluation examples or groups and recomputing the metric.

\subsection{Fixed-Claim Metrics}

For same-claim diagnostics, reference-label probability drop is:
\[
\Delta_i =
P_{\theta}(y_{\mathrm{ref}}\mid c_i,E_i^{\mathrm{ref}})
-
P_{\theta}(y_{\mathrm{ref}}\mid c_i,E_i^{\mathrm{hard}}).
\]
Probability-drop success is the fraction of pairs with $\Delta_i>0$. Strict swap success additionally requires the reference-evidence side to be predicted as the reference label and the insufficient-evidence side to be predicted as \textsc{NEI}.

\subsection{Prediction Coverage}

Prediction coverage is:
\[
\mathrm{coverage}=
\frac{n_{\mathrm{predicted}}}{n_{\mathrm{expected}}}.
\]
Rows with missing or duplicate predictions are not treated as complete.

\begin{table}[H]
\centering
\scriptsize
\begin{adjustbox}{max width=\columnwidth}
\begin{tabular}{lrrr}
\toprule
\textbf{Scope group} & \textbf{Expected} & \textbf{Predicted} & \textbf{Coverage} \\
\midrule
Fixed-claim reference side & 114 & 114 & 1.000 \\
Fixed-claim insufficient side & 114 & 114 & 1.000 \\
Same-document hard side & 114 & 114 & 1.000 \\
SciFact matrix regression & 885 & 885 & 1.000 \\
\bottomrule
\end{tabular}
\end{adjustbox}
\caption{Prediction coverage summary. Values summarize the repeated complete-coverage pattern across train variants and seeds.}
\label{tab:app-coverage-summary}
\end{table}

\subsection{Reporting Rules}

NEI-CAP follows these rules: report construction family for every \textsc{NEI} result; report Macro-F1 only when all three labels are meaningful; do not report Macro-F1 on one-class hard-\textsc{NEI} subsets; document prediction coverage before interpreting fixed-claim diagnostics; and mark FEVER/HoVer controls as bounded unless human validation is available.

%% file: appendices/M_Reproducibility_Release_and_Claim_Boundary_Checklist.tex
\section{Reproducibility, Release, and Claim Boundary Checklist}
\label{app:reproducibility}

This appendix records release readiness, the output-freeze policy, and paper-facing claim boundaries.

\subsection{Release Audit}

\begin{table}[H]
\centering
\scriptsize
\begin{adjustbox}{max width=\columnwidth}
\begin{tabular}{p{0.34\linewidth}ccp{0.38\linewidth}}
\toprule
\textbf{Check} & \textbf{Passed} & \textbf{Critical} & \textbf{Detail} \\
\midrule
Key reports exist & yes & yes & Experiment reports and summaries are present. \\
Paper tables exist and are non-empty & yes & yes & Paper-facing tables are generated. \\
Run registry entries exist & yes & yes & Core experiment entries are tracked. \\
Fixed-claim labels are preserved & yes & yes & Final fixed-claim/same-document labels path exists. \\
No LLM labels marked human & yes & yes & Reports use boundary language. \\
No candidate-only HoVer rows marked human-validated & yes & yes & HoVer marked human\_validated=no. \\
No Macro-F1 on one-class human subsets & yes & yes & One-class hard-\textsc{NEI} evaluations exclude Macro-F1. \\
Primary SciFact matrix primary outputs unchanged & yes & yes & Primary matrix results preserved unchanged. \\
External dependencies documented & yes & no & Upstream data dependencies documented. \\
Known limitations documented & yes & yes & Limitations recorded. \\
Unavailable cases documented & yes & no & Missing row-level case reasons tracked. \\
\bottomrule
\end{tabular}
\end{adjustbox}
\caption{NEI-CAP release audit checklist. Critical checks concern result traceability, label provenance, metric validity, and claim boundaries.}
\label{tab:app-release-audit}
\end{table}

\subsection{Primary Outputs and Traceability}

The primary evidence set includes the SciFact construction matrix, SciFact human-audit and hard-\textsc{NEI} evaluation, fixed-claim/same-document adjudicated labels, secondary-backbone robustness, FEVER subset controls, HoVer candidate missing-hop controls, and derived paper tables/figures. Each reported result should be traceable to a source dataset, construction family, split policy, model configuration, seed set, prediction artifact, and evaluation output.

\subsection{Release Package Contents}

The accompanying release will include construction manifests, group-disjoint split files, human-adjudicated labels, paper-facing prediction logs, evaluation scripts, artifact-audit scripts, and table-generation scripts. Upstream datasets are referenced and documented but not redistributed. Rows or case studies unavailable in primary outputs are documented rather than reconstructed from aggregate statistics.

\subsection{Claim Strength Matrix}

\begin{table}[H]
\centering
\scriptsize
\begin{adjustbox}{max width=\columnwidth}
\begin{tabular}{p{0.20\linewidth}p{0.14\linewidth}p{0.30\linewidth}p{0.26\linewidth}}
\toprule
\textbf{Claim} & \textbf{Strength} & \textbf{Allowed wording} & \textbf{Forbidden wording} \\
\midrule
NEI-CAP is construction-aware & Strong & Protocol for constructing, auditing, and stress-testing \textsc{NEI} evaluation. & Do not claim it solves fact verification. \\
SciFact \textsc{NEI} is construction-sensitive & Strong & Scores vary substantially across construction conditions. & Do not claim all errors have the same cause. \\
Easy \textsc{NEI} inflates performance & Moderate & Shortcut-like constructions can inflate apparent competence. & Do not claim every model uses only shortcuts. \\
Human validation confirms hard subsets & Strong & Audited BM25/cited and Fixed-claim/same-doc hard-\textsc{NEI} subsets are human-adjudicated. & Do not call LLM triage human validation. \\
Same-claim is diagnostic & Limited & Supports hard-side recognition with reference-side caveats. & Do not claim clean counterfactual evidence-use success. \\
HoVer is external control & Limited & Candidate-only multi-hop construction-sensitivity probe. & Do not call HoVer human-validated. \\
FEVER is external control & Limited & Non-toy subset-level shortcut control. & Do not call FEVER full-data validation. \\
Hard \textsc{NEI} is heterogeneous & Moderate & Different construction families stress different behaviors. & Do not collapse topic-unrelated cases into hard \textsc{NEI}. \\
Release is reproducible with limitations & Strong & Includes reproducibility instructions and claim boundaries. & Do not report toy, failed, or smoke outputs as paper-facing. \\
\bottomrule
\end{tabular}
\end{adjustbox}
\caption{Claim strength and wording boundaries.}
\label{tab:app-claim-strength}
\end{table}

\subsection{Final Release Audit}

The primary matrix results were frozen before the follow-up analyses and are unchanged, and no analysis created new human labels. The final audit covers split statistics, mixed training, multi-seed checks, the same-document artifact audit, and human-protocol documentation; it confirms no Macro-F1 on one-class hard subsets, no LLM-only human validation, and no candidate-only HoVer human validation. NEI-CAP is an audit and reporting protocol, not a blanket declaration that a dataset is invalid.

%% file: appendices/N_Decoder_SFT_Diagnostics.tex
\section{Decoder Zero-Shot and Controlled SFT Diagnostics}
\label{app:decoder-sft}

This appendix documents the decoder diagnostic summarized in Section~\ref{subsec:decoder-results}: a deterministic zero-shot evaluation of Qwen2.5-7B-Instruct and a controlled five-seed LoRA fine-tuning comparison in which only the training \textsc{NEI} realization changes. The question is whether fine-tuning changes a decoder verifier's sensitivity to the \textsc{NEI} construction used during training.

\subsection{Zero-Shot Evaluation and Class Bias}

The zero-shot evaluation uses deterministic conditional label scoring over the frozen SciFact construction tests (five constructions $\times$ 177 examples), the held-out human-adjudicated hard-\textsc{NEI} split ($n=54$), and the fixed-claim paired diagnostic ($n=114$ pairs). On the frozen core set, zero-shot Qwen reaches accuracy 0.775 and Macro-F1 0.725, with \textsc{Support}-F1 0.818, \textsc{Refute}-F1 0.532, and \textsc{NEI}-F1 0.824. Its easy-family mean \textsc{NEI}-F1 (over placeholder, random-irrelevant, and position-biased evaluation) is 0.845 and its BM25/cited hard mean is 0.793, so the zero-shot construction gap is modest. Human-hard recall is high (0.963), but the predicted \textsc{Support}/\textsc{Refute}/\textsc{NEI} rates are 45.4/9.3/45.3\% against a gold distribution of 42.9/22.6/34.5\%, so part of that recall comes from \textsc{NEI} overprediction and \textsc{Refute} underprediction.

Claim-only and shuffled-evidence controls caused the model to predict almost exclusively \textsc{NEI}, and we did not observe direct claim-only recovery of the original \textsc{Support}/\textsc{Refute} labels. These controls are diagnostic proxies: they neither prove contamination nor decontamination, and benchmark pretraining exposure remains unknown.

\subsection{Controlled SFT Design and Data Gates}

Across the three SFT regimes, only the training \textsc{NEI} evidence realization changes. The base model and revision, the \textsc{Support}/\textsc{Refute} multiset, claim groups, aligned \textsc{NEI} units, training size and label counts, the shared balanced development set, the checkpoint-selection metric, and the conditional-label evaluator are identical under explicit equality gates. After removing two aligned \textsc{NEI} groups that overlapped evaluation or shared-development assets, every regime trains on $n=831$ examples (774 claim groups; 343 \textsc{Support}, 194 \textsc{Refute}, 294 \textsc{NEI}).

\begin{table}[H]
\centering
\scriptsize
\begin{adjustbox}{max width=\columnwidth}
\begin{tabular}{lp{0.5\linewidth}r}
\toprule
\textbf{Regime} & \textbf{Training \textsc{NEI} realization} & \textbf{\textsc{NEI} $n$} \\
\midrule
Placeholder-only & Placeholder & 294 \\
Hard mixture & BM25 near-miss (148) + cited non-rationale (146) & 294 \\
Balanced-all & BM25 (51) + cited (59) + placeholder (60) + position-biased (58) + random irrelevant (66) & 294 \\
\bottomrule
\end{tabular}
\end{adjustbox}
\caption{Decoder SFT training regimes. Each aligned \textsc{NEI} unit contributes exactly one construction realization, so \textsc{Support}/\textsc{Refute} rows and \textsc{NEI} claim units are not multiplied across regimes.}
\label{tab:app-decoder-regimes}
\end{table}

All regimes share one byte-identical balanced development set ($n=179$; 166 groups; 89 \textsc{Support}, 31 \textsc{Refute}, 59 \textsc{NEI} spanning all five families). One global learning rate was selected from mean shared-development Macro-F1 across all three regimes; test, human-hard, and paired-diagnostic data were never used for learning-rate, checkpoint, or hyperparameter selection. Leakage gates all pass with zero intersections: train/development/core-test claim groups, normalized claim texts against all evaluation assets, normalized prompt contents across candidate \textsc{NEI} realizations, and source example IDs. Five seeds are reported per regime; the expansion beyond the three-seed primary evaluation followed a predeclared rule, triggered by placeholder-only seed variability above threshold, and test results never fed back into hyperparameters.

\subsection{Per-Regime Results}

\begin{table}[H]
\centering
\scriptsize
\begin{adjustbox}{max width=\columnwidth}
\begin{tabular}{lrrrrr}
\toprule
\textbf{Qwen regime} & \textbf{Easy \textsc{NEI}-F1} & \textbf{Hard \textsc{NEI}-F1} & \textbf{Easy--hard gap} & \textbf{Hum.\ recall} & \textbf{REF-F1} \\
\midrule
Zero-shot base & 0.845 & 0.793 & 0.052 & 0.963 & 0.532 \\
Placeholder-only SFT & 0.559 $\pm$ 0.192 & 0.054 $\pm$ 0.054 & 0.505 $\pm$ 0.140 & 0.011 & 0.645 \\
Hard-mixture SFT & 0.916 $\pm$ 0.014 & 0.834 $\pm$ 0.008 & 0.082 $\pm$ 0.010 & 0.900 & 0.784 \\
Balanced-all SFT & 0.953 $\pm$ 0.014 & 0.817 $\pm$ 0.018 & 0.135 $\pm$ 0.019 & 0.856 & 0.795 \\
\bottomrule
\end{tabular}
\end{adjustbox}
\caption{Full decoder regime results. Easy \textsc{NEI}-F1 is the mean over placeholder, random-irrelevant, and position-biased evaluation and is not the matched-placeholder score: under placeholder-only SFT, matched-placeholder \textsc{NEI}-F1 remains 1.000 in every seed even though the broader easy-family mean is 0.559, because the placeholder-specific rule also degrades the other easy families. SFT rows are descriptive five-seed means with sample standard deviations.}
\label{tab:app-decoder-results}
\end{table}

All five placeholder-SFT training-induced gap deltas are positive, and their paired group-bootstrap 95\% intervals~\citep{koehn2004statistical} exclude zero. The fixed-claim strict-pair success rate shows the same qualitative training dependence: 0.535 for the zero-shot base, 0.126 after placeholder-only SFT, and 0.653/0.654 after hard-mixture/balanced SFT. The balanced regime is not uniformly better than the zero-shot base: its human-hard recall is lower (0.856 vs.\ 0.963). It does, however, move the predicted label distribution from 45.4/9.3/45.3\% to 43.0/22.7/34.3\%, close to the gold 42.9/22.6/34.5\%, and raises \textsc{Refute}-F1 from 0.532 to 0.795.

\subsection{Llama Exclusion Gate}

Llama-3.1-8B-Instruct failed the predeclared fully permuted label-order stability gates: its maximum construction \textsc{NEI}-F1 delta across label orders was 0.222 under corrected generation (Qwen: 0.032) and 0.164 under conditional label scoring, with maximum predicted-label-rate deltas of 0.254 and 0.243 (Qwen: 0.040). Llama was therefore not fine-tuned; its zero-shot behavior is reported only as an exploratory prompt-sensitivity observation.

\subsection{Boundary}

The bounded claim is that training-induced construction sensitivity occurs in Qwen2.5-7B under the frozen SciFact setup, and that construction-diverse SFT is a practical mitigation. The evidence does not support a universal decoder collapse, a claim that other LLMs behave like Qwen, decontamination, production readiness, or LLM outputs as human validation.